\documentclass[preprint,12pt]{elsarticle}

\usepackage{cite}
\usepackage{amsmath,amssymb,amsfonts}
\usepackage{algorithmic}
\usepackage{graphicx}
\usepackage{algorithm,algorithmic}
\usepackage{hyperref}
\usepackage{textcomp}
\usepackage{microtype}
\usepackage{palatino} 
\usepackage{array} 
\usepackage{multirow}
\usepackage{graphicx}
\usepackage{algorithm}
\usepackage{csquotes}
\usepackage[table,xcdraw]{xcolor}  
\usepackage{mdframed} 
\usepackage{amsmath, amssymb} 
\usepackage{listings} 
\usepackage{microtype}
\usepackage{multirow}
\usepackage[normalem]{ulem}

\usepackage[normalem]{ulem}
\useunder{\uline}{\ul}{}

\usepackage{booktabs}
\usepackage{array}

\usepackage{tabularx}

\usepackage{fullpage}

\definecolor{codegreen}{rgb}{0,0.6,0}
\definecolor{codegray2}{rgb}{0.5,0.5,0.5}
\definecolor{codegray}{rgb}{0.9569, 0.9647, 0.9647}
\definecolor{yellow}{rgb}{0.9765, 0.9059, 0.6235}

\definecolor{myred}{rgb}{0.6627, 0.1961, 0.1490}
\definecolor{codepurple}{rgb}{0.58,0,0.82}
\definecolor{backcolour}{rgb}{0.95,0.95,0.92}
\definecolor{blue}{rgb}{0,0,0}

\lstdefinestyle{mystyle}{
    backgroundcolor=\color{backcolour},   
    commentstyle=\color{codegreen},
    keywordstyle=\color{magenta},
    numberstyle=\tiny\color{codegray},
    stringstyle=\color{codepurple},
    basicstyle=\tiny\ttfamily,
    breakatwhitespace=false,         
    breaklines=true,                 
    captionpos=b,                    
    keepspaces=true,                 
    numbers=left,                    
    numbersep=5pt,                  
    showspaces=false,                
    showstringspaces=false,
    showtabs=false,                  
    tabsize=2
}

\newmdenv[backgroundcolor=backcolour, linecolor=white, leftmargin=10, rightmargin=10, innerleftmargin=15, innerrightmargin=15, innertopmargin=10, innerbottommargin=10]{codeframe}

\renewcommand{\arraystretch}{1.2}

\usepackage{lineno}

\journal{Computers in Biology and Medicine}

\begin{document}

\begin{frontmatter}



\title{DALL-M: Context-Aware Clinical Data Augmentation with \textcolor{blue}{Large Language Models}} 

\author[label1]{Chihcheng Hsieh} 
\author[label3,label6,label2]{Catarina Moreira} 
\author[label4]{Isabel Blanco Nobre} 
\author[label4]{Sandra Costa Sousa} 
\author[label1]{Chun Ouyang} 
\author[label2]{Margot Brereton} 
\author[label5,label6]{Joaquim Jorge} 
\author[label5]{Jacinto C. Nascimento} 

\affiliation[label1]{organization={School of Information Systems, Queensland University of Technology},
            addressline={2 George Street}, 
            city={Brisbane},
            postcode={4000}, 
            state={QLD},
            country={Australia}}

\affiliation[label2]{organization={School of Computer Science, Queensland University of Technology},
            addressline={2 George Street}, 
            city={Brisbane},
            postcode={4000}, 
            state={QLD},
            country={Australia}}

\affiliation[label3]{organization={Data Science Institute, University of Technology Sydney},
            addressline={61 Broadway, Ultimo}, 
            city={Sydney},
            postcode={2007}, 
            state={NSW},
            country={Australia}}

\affiliation[label4]{organization={Imagiology Department, Grupo Lusiadas},
            addressline={Rua Abílio Mendes 12}, 
            city={Lisboa},
            postcode={1500-458}, 
            country={Portugal}}

\affiliation[label5]{organization={Instituto Superior Tecnico, Universidade de Lisboa},
            addressline={Av. Rovisco Pais 1}, 
            city={Lisboa},
            postcode={1049-001}, 
            country={Portugal}}

\affiliation[label6]{organization={INESC-ID},
            addressline={Av. Rovisco Pais 1}, 
            city={Lisboa},
            postcode={1049-001}, 
            country={Portugal}}
            
\begin{abstract}

X-ray images are vital in medical diagnostics, but their effectiveness is limited without clinical context. Radiologists often find chest X-rays insufficient for diagnosing underlying diseases, necessitating the integration of structured clinical features with radiology reports.

\textcolor{blue}{To address this, we introduce DALL-M, a novel framework that enhances clinical datasets by generating contextual synthetic data. DALL-M augments structured patient data, including vital signs (e.g., heart rate, oxygen saturation), radiology findings (e.g., lesion presence), and demographic factors. It integrates this tabular data with contextual knowledge extracted from radiology reports and domain-specific resources (e.g., Radiopaedia, Wikipedia), ensuring clinical consistency and reliability.}

\textcolor{blue}{DALL-M follows a three-phase process: (i) clinical context storage, (ii) expert query generation, and (iii) context-aware feature augmentation. Using large language models (LLMs), it generates both contextual synthetic values for existing clinical features and entirely new, clinically relevant features.}

\textcolor{blue}{Applied to 799 cases from the MIMIC-IV dataset, DALL-M expanded the original 9 clinical features to 91. Empirical validation with machine learning models—including Decision Trees, Random Forests, XGBoost, and TabNET—demonstrated a 16.5\% improvement in F1 score and a 25\% increase in Precision and Recall.}

\textcolor{blue}{DALL-M bridges an important gap in clinical data augmentation by preserving data integrity while enhancing predictive modeling in healthcare. Our results show that integrating LLM-generated synthetic features significantly improves model performance, making DALL-M a scalable and practical approach for AI-driven medical diagnostics.}

\end{abstract}


\end{frontmatter}


\section{Introduction}\label{sec:introduction}

Data augmentation is a fundamental technique in deep learning designed to enhance the diversity and robustness of training datasets. In image processing, techniques such as translation, rotation, scaling, and flipping introduce new patterns and variations, helping models generalize to unseen data. Similarly, in natural language processing (NLP), methods like word masking, substitution, and reordering improve a model’s ability to capture contextual relationships. \textcolor{blue}{However, augmenting structured tabular data in clinical settings—such as patient vitals, laboratory values, and imaging-derived features—presents unique challenges~\citep{liu2024survey, borisov2022language}. Unlike conventional augmentation techniques, medical data requires preserving the relationships between features (e.g., ensuring consistency between oxygen saturation and respiratory rate) to maintain clinical validity. Our approach ensures these relationships remain intact by integrating structured patient data with contextual information from radiology reports and trusted clinical knowledge bases.}

Clinical tabular data is highly structured and context-sensitive. Each data point, such as patient demographics, medical conditions, or test results, forms part of a more extensive, interdependent system. \textcolor{blue}{Modifying these features without preserving their inherent relationships can introduce inconsistencies that compromise clinical validity.~\citep{theodorou2023synthesize, bisercic2023interpretable}}. For instance, \textcolor{blue}{if a patient's blood pressure is altered without adjusting related factors such as age or comorbidities, the dataset may no longer align with real-world medical patterns.} Such discrepancies can mislead machine learning models, \textcolor{blue}{resulting in incorrect risk assessments or inaccurate diagnostic predictions~\citep{chen2021synthetic,gonzales2023synthetic}.}

\textcolor{blue}{Unlike traditional augmentation methods that introduce randomness, our proposed framework ensures that newly generated data maintains medical integrity, preserving the logical and clinical consistency of real patient data. This is achieved through domain-aware augmentation techniques that incorporate structured medical knowledge.}

\subsection{Research Gap}

Traditional augmentation methods, such as random shuffling or noise injection, often disrupt the complex relationships between clinical features, undermining the predictive power of machine learning models. This is especially problematic in healthcare, where patient safety and diagnostic accuracy are paramount. Introducing noise into clinical metrics—like blood pressure or glucose levels—may simulate pathological conditions that do not exist, resulting in misleading patterns during model training. Such distortions pose significant risks, potentially leading to diagnostic errors or inappropriate treatment recommendations, directly affecting patient outcomes.

\textcolor{blue}{Moreover, conventional augmentation methods fail to capture the interdependencies between medical features. For instance, artificially increasing a patient’s heart rate without adjusting oxygen saturation or respiratory rate can introduce clinically implausible cases. Similarly, duplicating data through synthetic oversampling does not add new knowledge, whereas DALL-M derives medically valid features from clinical context using LLMs and retrieval-augmented knowledge.}

Despite the availability of large medical datasets such as the UK Biobank \citep{uk_biobank}, clinical practice often involves working with smaller, less diverse datasets. The high cost of collecting and curating medical data exacerbates this issue, limiting many institutions' ability to develop robust deep learning models. Deep learning thrives on vast amounts of diverse data, and when clinical datasets are limited in size or variety, the ability to train models effectively is hindered. These limitations highlight a critical gap in how we approach data augmentation for clinical tabular data.

There is an urgent need for novel data augmentation techniques that can preserve the integrity and context of clinical datasets while enabling advanced machine learning models, even in data-constrained environments.

\subsection{Proposed Approach}

To address these challenges, we propose a novel data augmentation framework that leverages the reasoning capabilities of Large Language Models (LLMs). While LLMs are traditionally applied in natural language tasks, their ability to infer patterns across structured, heterogeneous data makes them promising candidates for clinical data augmentation. LLMs can be fine-tuned to understand domain-specific knowledge, enabling them to generate new data points that are logically consistent with existing clinical information.

In this context, we define "inferable" features as those that maintain the relationships between clinical variables, adhering to established medical knowledge and logic. For example, an LLM trained on clinical data can infer additional symptoms or disease progressions that align with a patient's existing conditions, generating new features that respect clinical constraints. However, fine-tuning LLMs for clinical tasks is challenging: the computational demands, potential biases inherited from pre-trained models, and the costs of frequent updates present significant obstacles.

Our approach mitigates these concerns by anchoring the LLM outputs to trusted clinical sources (such as Radiopedia) and rigorously validating the generated data through a classification task and expert reviews. This ensures that the augmented data is clinically valid and useful for downstream machine learning applications.

\subsection{Contributions}
This paper addresses the key research question: \textit{How can Large Language Models effectively augment clinical datasets with inferable, contextually relevant features while preserving clinical integrity?} We answer this by introducing a novel framework for structured data augmentation, named DALL-M, which generates clinically coherent features through a three-phase process: clinical context storage, expert query generation, and context-aware feature augmentation.

Our contributions are as follows:
\begin{enumerate}
    \item \textbf{Novel framework for contextually relevant augmentation (DALL-M)}: We introduce a framework that leverages LLMs to generate new clinical features, overcoming the limitations of traditional methods such as random shuffling or noise injection. Unlike these conventional methods, DALL-M ensures that the newly generated features are logically consistent with existing clinical data, maintaining data integrity.


    \item \textbf{Empirical validation of LLM-generated values for clinical features}: We present empirical evidence that DALL-M can generate clinically relevant and contextually accurate values for existing clinical features using LLMs. This validation shows that LLMs can synthesize patient-specific values that are statistically similar to real clinical data, thereby preserving the integrity and trustworthiness of the augmented dataset. By comparing LLM-generated values to ground-truth clinical measurements, we demonstrate that DALL-M outperforms traditional augmentation methods, which often introduce unrealistic or clinically inconsistent patterns.

    \item \textbf{Integration of LLM-augmented datasets with machine learning models}: We demonstrate how DALL-M enhances the performance of various machine learning models, including Decision Trees, Random Forests, XGBoost, and TabNET. By showing how contextually augmented features improve model robustness and accuracy, we address the challenge of applying deep learning to limited clinical data.

    \item \textbf{Anchoring to trusted clinical sources}: To mitigate the risk of hallucinations—where models generate fabricated or inaccurate clinical data—we anchor the LLM to trusted sources like Radiopaedia. This ensures the generated features are grounded in verified medical knowledge, safeguarding the clinical validity of the augmented data.

\end{enumerate}

Our approach enhances model performance while preserving patient-centric data relationships by ensuring that the LLM-augmented clinical features are inferable, contextually relevant, and grounded in clinical logic. This method bridges the gap between limited clinical data and the need for robust machine learning models, offering a reliable, scalable solution for clinical data augmentation.

\section{Related work}

\textcolor{blue}{
The augmentation of tabular data plays a crucial role in improving the robustness and generalizability of machine learning models, particularly in clinical applications where data availability is often limited. Traditional data augmentation methods for structured datasets have primarily relied on retrieval-based or generation-based approaches, each with distinct advantages and limitations. While retrieval-based techniques enrich datasets by incorporating external information, they are constrained by data availability and scalability issues. Conversely, generation-based methods leverage probabilistic models and deep learning to synthesize new data points but often struggle with overfitting and maintaining data integrity. In this section, we review conventional tabular data augmentation techniques, highlight their limitations, and position our proposed framework, DALL-M, within the broader landscape of machine learning-driven clinical data augmentation.}

\subsection{Conventional Approaches to Tabular Data Augmentation (TDA)}

\textcolor{blue}{
Conventional tabular data augmentation (TDA) can be roughly classified into retrieval-based or generation-based approaches \citep{1}. The first category enhances the original table with realistic data sourced from table pools, while the second generates synthetic data without relying on external sources. Generation-based methods leverage generative models such as BERT \citep{2}, T5 \citep{3}, large language models like ChatGPT \citep{4} and LLaMA \citep{5}, as well as deep learning architectures including variational autoencoders (VAE) \citep{6}, GANs \citep{7}, and diffusion models \citep{8}}.

\textcolor{blue}{
Retrieval-based approaches can be further divided into four main subcategories. The first is entity-based augmentation at the row level \citep{9,10}, which includes statistical \citep{9,11,12}, knowledge base-based \citep{13,14,15,16,17}, graph-based \citep{18,19,20}, and pre-trained language model (PLM)-based \citep{21,22} techniques. The second subcategory, schema augmentation at the column level, involves value-based \citep{23,24}, semantic-based \citep{25,26,27,28,29}, and structure-based \citep{30,31,32} joins. The third subcategory, cell completion at the cell level, uses attribute name \citep{15,18,33}, entity ID or name \citep{15}, and cell value \citep{15,18,33-35} completion methods. Finally, table integration at the table level consists of compositional \citep{15,18,33} and direct \citep{36,37} integration strategies.}

\textcolor{blue}{
Generation-based approaches are categorized into four sub-classes. Record generation techniques can be either distribution-preserving \citep{38,39,40,41,42,43} or designed for class-imbalance correction \citep{44,45,46,47}. Feature construction techniques are further classified into explicit \citep{48,49,50} and implicit \citep{51,52} approaches. Cell imputation strategies involve statistical methods \citep{53,54} or deep learning-based methods such as VAEs \citep{6}, GANs \citep{7}, and diffusion models \citep{8}. Lastly, table synthesis aims to generate entire synthetic tables \citep{55}.
}

\textcolor{blue}{
Despite the advancements in TDA, limitations persist. Retrieval-based methods extract related tables from table pools to perform data augmentation, enriching the original table with real external data. However, this approach depends on the availability of relevant information and requires preprocessing and indexing a potentially vast number of tables, raising scalability concerns. Additionally, the scarcity of labeled data in large-scale table pools suggests self-supervised approaches as a promising future direction. Furthermore, retrieval-based methods struggle with robust generalization. On the other hand, generation-based TDA synthesizes artificial data specifically for augmentation. Unlike retrieval-based strategies, these approaches do not require external data sources, eliminating preprocessing and indexing overhead. However, generation-based methods are often over-parameterized, leading to overfitting when working with small tables. Additionally, they lack interpretability and may produce hallucinated outputs, reducing the reliability of the generated data.}

\textcolor{blue}{
Our approach falls within the generation-based category, sharing similar concerns. However, the novelty of DALL-M lies in its context-aware feature generation, which leverages retrieval-augmented generation (RAG) techniques to enhance the clinical validity of synthetic data. Unlike conventional generation-based approaches that rely solely on statistical distributions or deep generative models, DALL-M incorporates structured domain knowledge from trusted clinical sources such as Radiopaedia and Wikipedia to inform and constrain the augmentation process.}

\textcolor{blue}{
DALL-M explicitly models the dependencies between clinical features, ensuring that generated synthetic data maintains medical consistency, such as heart rate variations remaining aligned with oxygen saturation and respiratory rate. This distinguishes our framework from traditional generative models, which often introduce inconsistencies by treating features independently. Another key advancement is DALL-M's ability to generate entirely new, clinically relevant features guided by expert-informed query generation. Instead of augmenting datasets with random perturbations, our method synthesizes additional patient attributes, enriching the dataset while preserving its interpretability. This mitigates the hallucination risk associated with large language models by anchoring feature generation to validated medical knowledge.}

\textcolor{blue}{
Additionally, DALL-M is specifically designed for tabular data in clinical settings, where structured dependencies must be maintained. Many existing generative approaches focus on image and text data, making direct adaptation to clinical tabular datasets challenging. By integrating domain-specific retrieval, structured query generation, and feature augmentation, our method ensures that synthetic data is both realistic and clinically meaningful.}

\subsection{LLM-Based Approaches for Data Augmentation}

Medical Large Language Models (Med-LLMs) are increasingly used in clinical workflows for tasks such as improving medical knowledge understanding, enhancing diagnostic accuracy, and providing personalized treatment recommendations. These capabilities contribute to precise decision-making, better patient care, and improved treatment outcomes. Several Med-LLMs, including ChiMed-GPT \citep{tian2023chimed}, MedicalGPT \citep{Xu2023}, HuatuoGPT-II \citep{chen2023huatuogpt}, and ChatMed \citep{zhu2023chatmed}, have gained significant attention in biomedical research. Existing studies can be categorized into two main areas \citep{liu2024survey}: current Med-LLMs in the medical field and their role in supporting clinical and patient treatment.

Several existing Med-LLM methodologies have emerged. HuatuoGPT \citep{chen2023huatuogpt} utilizes reinforcement learning to align LLMs with real-world data. ClinicalT5 \citep{lu2022clinicalt5} adapts general LLMs for clinical text by pre-training on MIMIC-III data. ClinicalGPT \citep{wang2023clinicalgpt} integrates medical records, domain-specific knowledge, and multi-round consultations. ChiMed-GPT \citep{tian2023chimed} trains on domain-specific knowledge, while BioGPT \citep{Luo2022} enhances biomedical text generation capabilities. Other models such as PubMedBERT \citep{gu2021domain}, GatorTron \citep{yang2022large}, and Med-PaLM \citep{singhal2023large} focus on various aspects of clinical data processing, including electronic health records and multimodal medical analysis.

Despite the broad applications of Med-LLMs, this study proposes utilizing them in a novel way. Previous studies such as GReaT \citep{borisov2022language}, HALO \citep{theodorou2023synthesize}, and TEMED-LLM \citep{bisercic2023interpretable} have explored Med-LLMs for tabular data, but none have specifically tackled clinical data augmentation. DALL-M fills this gap by addressing how to generate realistic clinical features and infer new attributes. Our methodology integrates domain experts in evaluating generated features, ensuring that synthetic clinical features are meaningful and valuable for machine learning applications in medical diagnostics.

\textcolor{blue}{Table \ref{tab:lit_rev} provides a comparative overview of existing LLM-based approaches for medical data augmentation, highlighting their key focus areas, limitations, and how DALL-M addresses these gaps by generating new clinically relevant features and enhancing tabular data integration.} 

\textcolor{blue}{In summary, DALL-M pioneers LLM-driven tabular data augmentation by combining retrieval-augmented contextualization, structured knowledge integration, and expert-guided feature generation—capabilities absent in prior methods. Unlike traditional augmentation techniques that rely on random permutations or noise injection, DALL-M ensures clinical validity by dynamically retrieving and synthesizing multi-source medical knowledge (e.g., Radiopaedia, Wikipedia). It generates entirely new, context-aware clinical features, bridging the gap between text-based LLMs and structured data augmentation. By incorporating expert-driven constraints, DALL-M mitigates biases, hallucinations, and inconsistencies, leading to more interpretable, reliable, and clinically relevant synthetic data, ultimately improving AI-driven healthcare predictions and decision-making.}

\begin{table}[]
\resizebox{\columnwidth}{!}{
\begin{tabular}{|l|l|l|l|}
\hline
\textbf{Approach} &
  \textbf{Key Focus} &
  \textbf{Limitations} &
  \textbf{Role of DALL-M} \\ \hline
\multirow{3}{*}{\begin{tabular}[c]{@{}l@{}}ChiMed-GPT \citep{tian2023chimed}, \\ MedicalGPT \citep{Xu2023}, \\ HuatuoGPT-II \citep{chen2023huatuogpt}, \\ ChatMed \citep{zhu2023chatmed}\end{tabular}} &
  \begin{tabular}[c]{@{}l@{}}Adapt general-domain LLMs \\ for clinical tasks\end{tabular} &
  \begin{tabular}[c]{@{}l@{}}Primarily focused on text-based reasoning \\ (e.g., question answer tasks)\end{tabular} &
  \begin{tabular}[c]{@{}l@{}}Leverages LLMs in a novel way to enrich \\ \textit{tabular} data with new features and \\ synthetic values\end{tabular} \\ \cline{2-4} 
 &
  \begin{tabular}[c]{@{}l@{}}Harness large-scale medical \\ text corpora\end{tabular} &
  \begin{tabular}[c]{@{}l@{}}Less attention to generating entirely new \\ clinical features\end{tabular} &
  \begin{tabular}[c]{@{}l@{}}Goes beyond question-answering by creating \\ entirely new features relevant to specific \\ radiology contexts\end{tabular} \\ \cline{2-4} 
 &
  \begin{tabular}[c]{@{}l@{}}Provide domain-specific \\ responses (e.g., diagnoses)\end{tabular} &
  Limited exploration of tabular data &
  Focuses on text and tabular data \\ \hline
\multirow{2}{*}{ClinicalT5 \citep{lu2022clinicalt5}} &
  \begin{tabular}[c]{@{}l@{}}Fine-tuned T5 for clinical text tasks\\  (e.g., summarizing patient records)\end{tabular} &
  \begin{tabular}[c]{@{}l@{}}Focus on generating or refining text-based \\ outputs, not specialized in augmenting \\ structured (tabular) clinical data with new \\ features\end{tabular} &
  \begin{tabular}[c]{@{}l@{}}DALL-M introduces a three-phase pipeline \\ that leverages multiple data sources to create \\ augmented tabular data\end{tabular} \\ \cline{2-4} 
 &
  \begin{tabular}[c]{@{}l@{}}Shows improved performance in \\ typical NLP tasks such as entity \\ recognition, summarization\end{tabular} &
  \begin{tabular}[c]{@{}l@{}}Does not explore advanced context \\ integration (e.g., from multiple sources \\ like Radiopaedia, Wikipedia)\end{tabular} &
  \begin{tabular}[c]{@{}l@{}}Adds clinically relevant features rather than \\ just refining text\end{tabular} \\ \hline
\multirow{2}{*}{ClinicalGPT \citep{wang2023clinicalgpt}} &
  \begin{tabular}[c]{@{}l@{}}Integrates various info sources \\ (medical records, domain knowledge, \\ dialogues)\end{tabular} &
  \begin{tabular}[c]{@{}l@{}}Emphasis on multi-round consultations \\ and EHR text retrieval\end{tabular} &
  \begin{tabular}[c]{@{}l@{}}DALL-M not only consolidates domain \\ knowledge from multiple sources but also \\ generates new, context-aware \\ \emph{tabular} features\end{tabular} \\ \cline{2-4} 
 &
  \begin{tabular}[c]{@{}l@{}}Enables context-rich responses in \\ interactive clinical settings\end{tabular} &
  \begin{tabular}[c]{@{}l@{}}No explicit mechanism for structured \\ tabular feature augmentation or generation \\ of new features\end{tabular} &
  \begin{tabular}[c]{@{}l@{}}Uses a retrieval-augmented pipeline to \\ ensure alignment of new features with \\ established clinical knowledge\end{tabular} \\ \hline
\multirow{2}{*}{\begin{tabular}[c]{@{}l@{}}BioGPT \citep{Luo2022} \\ PubMedBERT \citep{gu2021domain} \\ GatorTron \citep{yang2022large}\end{tabular}} &
  Domain-trained LLMs &
  \begin{tabular}[c]{@{}l@{}}Specialized for textual data (biomedical \\ corpora, EHR text) with minimal focus on \\ synthetic feature generation in numerical \\ or tabular formats\end{tabular} &
  \begin{tabular}[c]{@{}l@{}}DALL-M embeds new numeric or categorical \\ features into existing tabular records, \\ preserving consistency and context\end{tabular} \\ \cline{2-4} 
 &
  \begin{tabular}[c]{@{}l@{}}Enhanced accuracy in medical text \\ classification, knowledge extraction, etc\end{tabular} &
  \begin{tabular}[c]{@{}l@{}}Do not address missing data or new feature \\ creation in a robust data-augmentation setting\end{tabular} &
  \begin{tabular}[c]{@{}l@{}}Addresses missing data scenarios and introduces \\ novel features (e.g., additional vital stats, \\ symptom descriptors)\end{tabular} \\ \hline
\multirow{3}{*}{\begin{tabular}[c]{@{}l@{}}Med-PaLM \citep{singhal2023large}\\ MedAlpaca \citep{han2023medalpaca}\\ LLaVA-Med \citep{li2024llava}\end{tabular}} &
  \begin{tabular}[c]{@{}l@{}}Employ large-scale domain adaptation \\ of general LLMs\end{tabular} &
  \begin{tabular}[c]{@{}l@{}}Focus primarily on question-answer or generative\\  tasks (report summarization, explanations)\end{tabular} &
  \begin{tabular}[c]{@{}l@{}}DALL-M explicitly tackles the augmentation \\ of \emph{tabular} clinical data with entirely new \\ features (not just text or imaging commentary)\end{tabular} \\ \cline{2-4} 
 &
  Incorporate specialized medical data &
  \multirow{2}{*}{\begin{tabular}[c]{@{}l@{}}Limited exploration of how to augment \emph{structured,}\\ \emph{tabular} clinical data with newly \\ inferred features\end{tabular}} &
  \multirow{2}{*}{\begin{tabular}[c]{@{}l@{}}Uses multi-modal and multi-source knowledge \\ (X-rays + textual reports + Radiopaedia + Wikipedia) \\ for robust feature generation\end{tabular}} \\ \cline{2-2}
 &
  \begin{tabular}[c]{@{}l@{}}Provide multi-modal understanding \\ (images + text)\end{tabular} &
   &
   \\ \hline
\multirow{2}{*}{\begin{tabular}[c]{@{}l@{}}TDA \\ ( Noise Injections, \\ Data Permut.)\end{tabular}} &
  Simple to implement for numeric data &
  \begin{tabular}[c]{@{}l@{}}Often breaks clinical consistency (random \\ noise can create medically implausible \\ combinations)\end{tabular} &
  \begin{tabular}[c]{@{}l@{}}DALL-M preserves clinical logic by integrating \\ domain knowledge from multiple sources and \\ structured expert queries\end{tabular} \\ \cline{2-4} 
 &
  \begin{tabular}[c]{@{}l@{}}Can increase dataset size quickly \\ for ML model training\end{tabular} &
  \begin{tabular}[c]{@{}l@{}}Does not leverage domain knowledge to ensure \\ realism or generate completely new features\end{tabular} &
  \begin{tabular}[c]{@{}l@{}}-Ensures newly generated or augmented features \\ remain medically plausible\end{tabular} \\ \hline
\multirow{4}{*}{\begin{tabular}[c]{@{}l@{}}DALL-M \\ ( Our approach)\end{tabular}} &
  \begin{tabular}[c]{@{}l@{}}Generates synthetic, context-relevant \\ numeric values for missing data\end{tabular} &
  \begin{tabular}[c]{@{}l@{}}Requires advanced retrieval-augmented methods \\ and curated knowledge sources\end{tabular} &
  \begin{tabular}[c]{@{}l@{}}Fills the gap of systematically augmenting tabular \\ data with \emph{both} realistic values for existing \\ columns and \emph{entirely new} clinically relevant \\ features\end{tabular} \\ \cline{2-4} 
 &
  \begin{tabular}[c]{@{}l@{}}Creates entirely new features based on \\ domain knowledge and retrieval\\ augmented LLM queries\end{tabular} &
  \begin{tabular}[c]{@{}l@{}}Dependent on LLM frameworks (e.g., GPT-4) \\ for best performance, which can be computationally \\ expensive\end{tabular} &
  \multirow{3}{*}{\begin{tabular}[c]{@{}l@{}}Provides robust, context-aware augmentation \\ that improves downstream ML performance and \\ addresses missing data issues\end{tabular}} \\ \cline{2-3}
 &
  \begin{tabular}[c]{@{}l@{}}Integrates multiple data sources \\ (e.g., Radiopaedia, Wikipedia) and \\ expert inputs\end{tabular} &
  \multirow{2}{*}{\begin{tabular}[c]{@{}l@{}}Ethical and privacy guidelines for synthetic data \\ must be carefully observed\end{tabular}} &
   \\ \cline{2-2}
 &
  \begin{tabular}[c]{@{}l@{}}Demonstrates efficacy across multiple\\ ML models (e.g., XGBoost, TabNet)\end{tabular} &
   &
   \\ \hline
\end{tabular}
}
\caption{\textcolor{blue}{Summary of LLM-Based Approaches for Data Augmentation.}}
\label{tab:lit_rev}
\end{table}

\section{DALL-M: Data Augmentation with LLMs}

\begin{figure*}[!h]
    \centering
    \includegraphics[width=\textwidth]{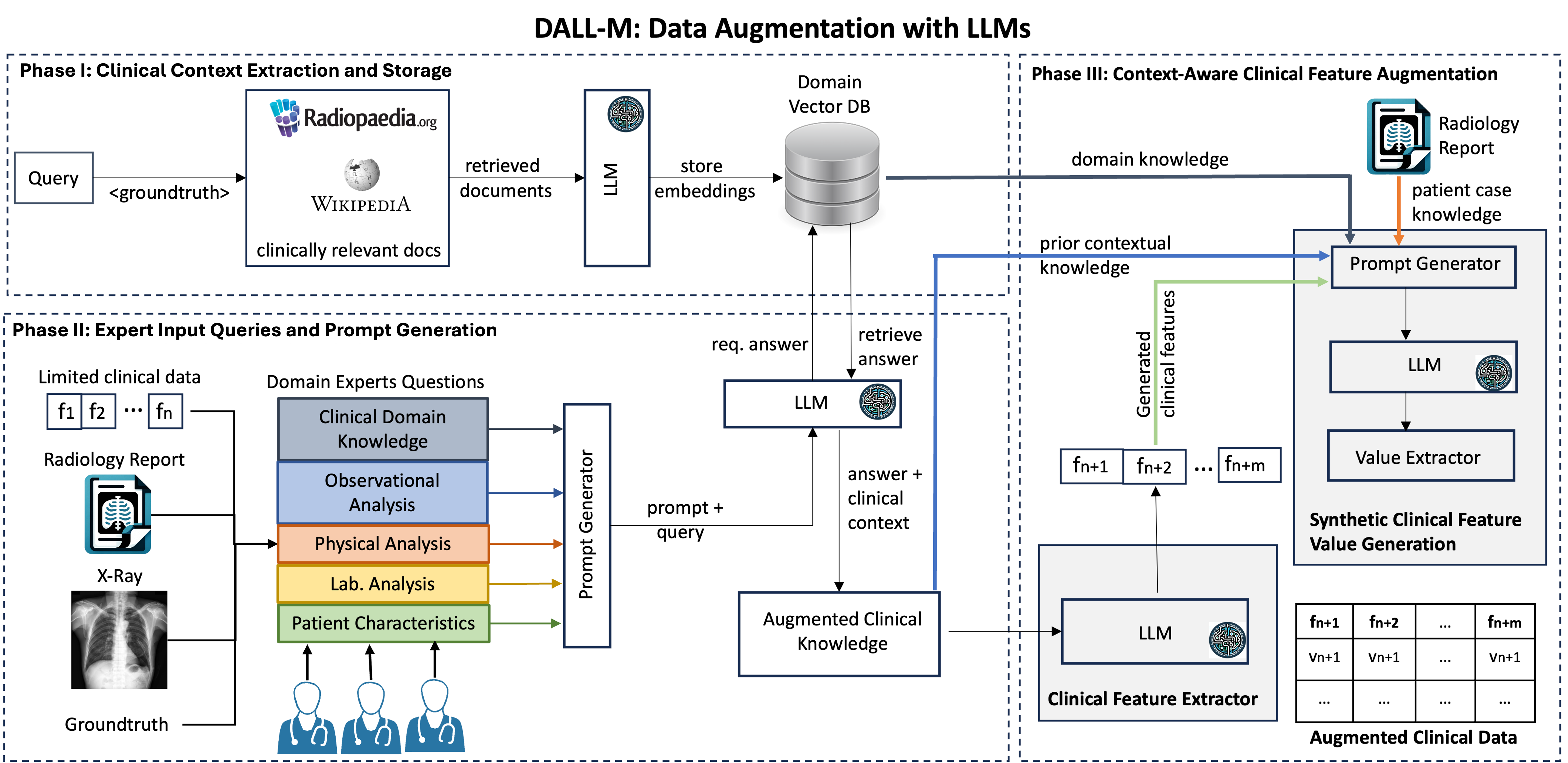}
    \caption{ Overview of the DALL-M framework for generating clinically relevant features using Large Language Models. The process consists of three main phases: (I) \textit{Clinical Context Extraction and Storage}, where patient-specific contexts and clinical relationships are captured and stored; (II) \textit{Expert Input Queries and Prompt Generation}, where medical experts provide contextual queries to guide the LLM; and (III) \textit{Context-Aware Clinical Feature Augmentation}, where new features are generated in alignment with the clinical context, ensuring that the augmented data remains consistent and clinically valid.
    }
    \label{fig:main}
\end{figure*}

The proposed framework, DALL-M, aims to enhance clinical datasets by generating new, contextually relevant features using LLMs. The framework consists of three key stages: (1) Clinical Context Extraction and Storage, (2) Expert Input Queries and Prompt Generation, and (3) Context-Aware Feature Augmentation. Together, these stages form a comprehensive data augmentation workflow, ranging from the extraction of clinical context to the generation of enhanced features, all while leveraging the analytical capabilities of LLMs. These stages are illustrated in Figure~\ref{fig:main}. The following sections provide detailed descriptions of each phase. \textcolor{blue}{To improve clarity, we introduce a running example throughout this section:}

\begin{quote}
\textcolor{blue}{Consider a patient whose chest X-ray shows evidence of pleural effusion. The initial dataset only contains three clinical features for this patient: age (65 years old), temperature (38.2°C), and oxygen saturation (92\%). Our goal is to augment this patient’s clinical data by generating additional relevant features while preserving medical integrity. DALL-M will first extract medical knowledge about pleural effusion (Phase I), use expert-informed queries to retrieve textual contents related to the patient's clinical context (Phase II), and finally generate new synthetic and clinically meaningful features (Phase III).}
\end{quote}

\subsection{Phase I: Clinical Context Extraction and Storage}\label{sec:phase1} 

The primary objective of this phase is the extraction and structured storage of clinically relevant context and patient-specific medical history. Beginning with a particular patient case, the pipeline systematically builds a comprehensive knowledge repository that encapsulates detailed clinical information about the patient's conditions. In this context, "conditions" refer specifically to lesions identified via ground truth labels from the REFLACX dataset \citep{bigolin2022reflacx}, a well-established radiological dataset that provides expert-annotated lesion labels as a foundation for subsequent knowledge extraction. \textcolor{blue}{A lesion in this context is an abnormality found in the x-ray, for instance, a lung nodule.}

The process begins by leveraging these lesion labels to query two authoritative and distinct online resources: Radiopaedia~\citep{radiopedia} and Wikipedia~\citep{wikipedia}. Radiopaedia, a leading online platform dedicated to radiology, provides a robust collection of peer-reviewed radiological cases, detailed articles, and reference materials, all aimed at enhancing clinical knowledge and decision-making in radiology. Wikipedia, on the other hand, is a vast, freely accessible encyclopedia that covers a wide range of topics, offering a more general contextual understanding that can complement specialized medical knowledge. Both resources are queried to establish a broad and detailed clinical foundation for each patient case, ensuring coverage of both domain-specific and general knowledge relevant to the identified lesions.

We employ the Retrieval Augmented Generation (RAG) framework to systematically process and organise this information. RAG is a hybrid methodology that integrates traditional information retrieval techniques with the generative power of LLMs. The framework comprises two components: a retriever model and a generator model. The retriever model is responsible for identifying and extracting relevant documents from external sources (in this case, Radiopaedia and Wikipedia) based on the lesion labels. These documents are then passed to the generator model, which processes the retrieved information and generates semantically coherent, contextually enriched text that aligns with the clinical requirements of the task. The integration of real-time information retrieval with LLMs ensures that the generated responses are not only accurate but also context-aware, providing a comprehensive clinical picture tailored to the patient's case. 

The output of the RAG process is then stored in a structured format to enable efficient querying and retrieval. After evaluating multiple database architectures, including relational and graph-based models, we determined that Neo4j (specifically utilizing its vector index capabilities) provides optimal performance for this task. Neo4j, a native graph database, is well-suited for capturing complex relationships between clinical entities, such as lesions, symptoms, and patient characteristics. The addition of vector indexing enhances this by enabling efficient similarity-based retrieval, ensuring semantically related information is easily accessible for subsequent phases. This configuration was meticulously applied to each of the five lesion labels extracted from the REFLACX dataset, yielding a domain-specific clinical knowledge graph that stores relevant clinical context and ensures rapid and accurate retrieval for downstream analysis.

\textcolor{blue}{One of the key advantages of using a RAG approach is that it anchors LLM responses to authoritative medical sources (e.g., Radiopaedia, Wikipedia). Unlike free-form LLM text generation, this grounding mechanism significantly reduces hallucinations, as all generated features and values are derived from verified medical knowledge. This ensures that synthetic data remains clinically coherent and minimally prone to incorrect or misleading outputs.}

\begin{quote}
\textcolor{blue}{\textbf{Running Example (Phase I):} The patient’s X-ray is labeled with a lesion “pleural effusion.” Our system retrieves detailed medical descriptions, causes, symptoms, and diagnostic indicators of pleural effusion from Radiopaedia and Wikipedia. This information is stored in a structured Neo4j database for later use. Key findings include: (1) Common causes such as heart failure and infection, (2) Typical symptoms like shortness of breath and chest pain, and (3) Diagnostic indicators such as blunting of the costophrenic angle.}
\end{quote}

To summarize, Phase I takes a specific clinical query (e.g., "pleural effusion"), retrieves the primary documents related to that query from Radiopaedia and Wikipedia and stores them in a database. This structured storage provides a strong foundation for clinical knowledge that will be used in later stages for data augmentation and feature generation.

\subsection{Phase II: Expert Input Queries and Prompt Generation} \label{sec:phase2}

After having the database populated with documents related to the user query, Phase II generates clinically relevant, contextually rich domain knowledge that mirrors radiologists' diagnostic process when assessing X-rays. 
 \textcolor{blue}{To ensure medical validity, we consulted two board-certified radiologists}, each with \textcolor{blue}{8 to 15 years of experience} in diagnostic imaging and chest X-ray interpretation. We conducted structured interviews with expert radiologists to understand the critical clinical questions that arise during their diagnostic workflow when they are presented with an X-Ray. They aimed to distil their expertise into focused queries guiding subsequent data augmentation and feature generation. \\

\noindent
\textit{Expert Queries}\\

\textcolor{blue}{Through these interviews, we transcribed and analyzed expert responses using thematic coding to extract common diagnostic reasoning patterns. This enabled us to systematically derive a structured query framework, ensuring that each lesion type is assessed through a standardized set of clinically relevant questions.}
These questions form a structured framework to guide data retrieval and augmentation, ensuring the clinical insights generated are comprehensive and aligned with radiological best practices.

\textcolor{blue}{To automate the application of these queries, we implemented a dynamic template-based approach where lesion-specific terms (e.g., "atelectasis" or "pleural effusion") are automatically inserted into predefined question formats. This ensures consistency and scalability in applying the expert queries across different patient cases.}

\begin{itemize}
    \item [] General Clinical Domain Knowledge:
    \begin{itemize}
            \item \textit{What are the most common symptoms associated with \{lesion\}?}
            \item \textit{What are the potential causes of \{lesion\}?}
    \end{itemize}
    These questions help establish a broad understanding of the lesion, providing a general clinical overview that can be applied across various cases. A lesion in this context is an abnormality found in the x-ray, for instance, a lung nodule.

    \item [] Observational Analysis:
    \begin{itemize}
            \item \textit{What are the patient’s symptoms that are relevant for \{lesion\}?}
    \end{itemize}
    This ensures that the patient's subjective experience is integrated into the clinical assessment, enriching the context of the diagnosis.

    \item [] Physical Examination and Imaging:
    \begin{itemize}
        \item \textit{What are the relevant clinical signs for the etiological diagnosis of \{lesion\}}\textcolor{blue}{?}

        \item \textit{What are the relevant clinical characteristics for the etiological diagnosis of \{lesion\}}\textcolor{blue}{?}
    \end{itemize}
    These queries capture the objective clinical observations and imaging findings that radiologists rely on to characterize and diagnose a lesion.

    \item [] Laboratory Data:
    \begin{itemize}
        \item What are the relevant laboratory data for the etiological diagnosis of \{lesion\}
    \end{itemize}
    These questions allow for integrating laboratory results and additional imaging tests into the diagnostic process.

    \item [] Patient Characteristics:
    \begin{itemize}
        \item What is the patient's personal, relevant history for the etiological diagnosis of \{lesion\}
    \end{itemize}
    These questions contextualize the clinical findings within the patient's broader medical history and demographic profile.
\end{itemize}

\textcolor{blue}{These structured queries serve as inputs to our Retrieval-Augmented Generation framework, which retrieves relevant clinical information from Radiopaedia and Wikipedia. This approach ensures that the augmentation process is informed by domain knowledge, dynamically adapting to different lesions while maintaining consistency with radiological best practices.}

\noindent
\textit{Prompt Generation and Clinical Information Retrieval}\\ 
Given these structured questions, the next step is to construct prompts that effectively query the contextual clinical database built in Phase I. Each question is translated into a prompt designed to elicit detailed responses from the LLMs through the RAG approach.

To achieve this, we apply a RAG process in the same two-step approach outlined in Phase I, wherein the retriever model identifies the most relevant documents from the domain-specific vector database, and the generator model synthesizes contextually enriched responses from the retrieved material. This approach ensures the generated outputs are grounded in authoritative clinical knowledge, tailored to the patient's specific case, and aligned with expert radiological practices.

The output of this phase is an Augmented Clinical Knowledge (ACK) corpus, a curated collection of clinical insights that corresponds to the answers to the expert queries. The ACK captures information about the patient's condition, including detailed explanations of the lesions, potential symptoms, and patient-specific factors (such as comorbidities or risk factors like obesity) that could influence the presentation of the lesion. This corpus enriches the clinical context for each patient case and provides critical input for the next phase, where new clinical features will be generated based on this augmented knowledge. This ensures that the subsequent data augmentation and feature generation processes are clinically meaningful and grounded in real-world radiological expertise.

\begin{quote}
\textcolor{blue}{\textbf{Running Example (Phase II):} For pleural effusion, the expert queries retrieve information such as: (1) Symptoms: Shortness of breath, chest pain, cough; (2) Possible causes: Heart failure, pneumonia, malignancy; (3) Relevant clinical features: Decreased breath sounds, dullness to percussion, jugular venous distension. These findings are stored in a structured format and fed into the next phase.}
\end{quote}

\subsection{Phase III: Context-Aware Feature Augmentation}\label{sec:phase3}

The primary objectives of this phase are twofold: (1) to identify and generate new clinically relevant features and (2) to assign meaningful values to these newly identified features. This process builds on the Augmented Clinical Knowledge corpus developed in Phase II, extracting novel clinical features that are contextually aligned with the patient’s condition. Extracting such features from unstructured textual data presents challenges, even with advanced techniques like named entity recognition, as medical text often includes nuanced and domain-specific terminology that is difficult to capture in its entirety.

To address these challenges, we leverage LLMs equipped with few-shot learning capabilities, allowing them to identify relevant features even with minimal labelled examples. The LLM can discern \textcolor{blue}{$n$} distinct features by examining the documents generated in Phase II, each feature representing a clinically important aspect of the patient’s condition. These features are not merely repetitions of existing variables but represent new data dimensions that provide deeper insights into the patient’s health status.

However, the mere identification of relevant clinical features is insufficient for a comprehensive augmentation of the clinical dataset. Each feature must be populated with corresponding values to ensure its practical utility. To achieve this, we revisit the RAG framework, crafting prompts that utilize multiple critical sources of information:

\begin{enumerate}
    \item \textbf{Prior contextual knowledge} derived from the ACK corpus in Phase II provides a comprehensive understanding of the patient’s condition.

    \item \textbf{Newly identified clinical features}, ensuring that the values generated directly relate to the novel attributes extracted in this phase.

    \item \textbf{The domain knowledge database established in Phase I} serves as a reference for validated medical information related to the lesion and patient.

    \item \textbf{The patient's radiology report and demographic details} offer personalized data for generating patient-specific feature values.
\end{enumerate}

By integrating these four sources of information, we generate precise and contextually relevant values for the newly identified features. For example, if a new feature related to lesion morphology is identified, values would be generated based on both the clinical descriptions in the radiology report and the broader domain knowledge available in the ACK corpus.

When processed by the LLM within the RAG framework, these prompts return a comprehensive list of features and their corresponding values. These results are then integrated into the original clinical dataset, culminating in an enriched and context-aware augmented dataset. This enriched dataset not only contains new clinically relevant features but also ensures that each feature is accompanied by accurate, personalized values, enhancing the dataset’s utility for subsequent analyses or predictive modeling.

\begin{quote}
\textcolor{blue}{\textbf{Running Example (Phase III):} DALL-M identifies that pleural effusion is commonly associated with the following new features: (1) Jugular Venous Distension (Yes/No), (2) Breath Sounds (Normal/Decreased/Absent), and (3) Chest Pain Severity (Mild/Moderate/Severe). The LLM assigns values based on the patient’s known attributes. Given that the patient is 65 years old, has a temperature of 38.2°C, and an oxygen saturation of 92\%, DALL-M predicts: Jugular Venous Distension = Yes, Breath Sounds = Decreased, and Chest Pain Severity = Moderate. These features are added to the dataset, enriching the patient’s clinical profile.}
\end{quote}

\textcolor{blue}{For more details, please refer to Appendix Figures \ref{fig:synthetic_generation} and \ref{fig:exp3_general}, where we outline how each component of the prompt interacts with the available data sources to yield high-quality, contextually appropriate results and present the general structure of our prompt formulation process and the methodology for assigning values.}

\textcolor{blue}{While expert input refines the selection of clinically relevant queries, it is important to emphasize that DALL-M does not depend solely on expert-designed prompting for consistency. Instead, its robustness is achieved through a structured, three-phase pipeline, where (i) retrieval-augmented context processing ensures factual grounding, and (ii) structured augmentation maintains logical consistency. Even without expert-driven prompt engineering, the system retrieves and synthesizes clinically coherent information, demonstrating that performance improvements are inherent to its design rather than reliant on manual tuning.}
\\

\section{Experimental Setup}

 This section details the experimental setup used to evaluate the effectiveness of the DALL-M framework. We describe the datasets utilized, the clinical features selected for evaluation, and the specific hypotheses tested through a series of experiments.

\subsection{Dataset}

In this work, we utilize instances from the MIMIC-IV dataset~\citep{Johnson2021MIMIC_IV}, comprising radiographs, radiology reports from chest X-rays, and clinical data. We gathered 799 patient cases from these datasets, with each instance accompanied by corresponding ground-truth information on clinical features, facilitating the evaluation of the generated synthetic data. The labels used for evaluation in Sect.~\ref{sec:experiment2} are extracted from the REFLACX dataset~\citep{bigolin2022reflacx}, annotated by five radiologists. In the dataset, we have the following clinical features: 
{\bf (1)}~\textbf{temperature} (in Fahrenheit degrees),  
{\bf (2)}~\textbf{heartrate}  (beats rate per minute), 
{\bf (3)}~\textbf{resprate} (breaths rate per minute), 
{\bf (4)}~\textbf{o2sat} (peripheral oxygen saturation as a percentage), 
{\bf (5)}~\textbf{sbp, dbp} (systolic and diastolic blood pressure, respectively, measured in millimeters of mercury (mmHg)). 
Additionally, the dataset incorporates ground truth labels indicating the presence/absence of lesions within the chest X-ray images of patients. We selected 
five distinct lesions 
for consideration: \textit{atelectasis}, \textit{consolidation}, \textit{enlarged cardiac silhouette}, \textit{pleural effusion}, and \textit{pleural abnormality}. The rationale behind selecting these specific lesions stem from their prevalence as the most frequently occurring conditions within the REFLACX dataset~\citep{Lanfredi2021REFLACX}.

\textcolor{blue}{Although MIMIC-IV~\citep{Johnson2021MIMIC_IV} contains over 40,000 patients, we curated a subset of 799 cases by ensuring that each sample contained all necessary modalities and labels for the study. The inclusion criteria required the presence of MIMIC-IV DICOM images, hospital module data (gender, age), emergency department triage features (such as vital signs), REFLACX lesion bounding box labels, and CheXpert labels \citep{irvin2019chexpert}. This intersection resulted in a final dataset of 799 cases.}

\textcolor{blue}{Additionally, while the five selected lesion classes are not perfectly balanced, we employed measures to mitigate potential biases, including data augmentation techniques and evaluation using Precision, Recall, and F1-score to ensure a fair assessment of model performance.}

\subsection{Experiments}

This section outlines the experiments designed to evaluate the proposed DALL-M framework. These experiments aim to validate two key hypotheses central to the framework's ability to generate contextual values for {\it existing} features and extend its knowledge to create {\it new} clinically relevant features.

\begin{itemize}
    \item \textbf{Hypothesis 1:} LLMs are capable of providing sufficient contextual information to generate accurate, patient-specific values for {\it existing} clinical features when provided with clinical patient reports.

    \item \textbf{Hypothesis 2:} LLMs can extend their knowledge to create entirely {\it new} clinically relevant features and their respective values, enhancing the original dataset.
\end{itemize}

To validate these hypotheses, we designed a series of experiments, each tailored to evaluate specific aspects of the LLMs' capabilities.\\

\noindent
\textbf{Validating Hypothesis 1: Can LLMs Generate Contextual Values for Existing Features? (see Section \ref{sec:generate-synthetic-values})}\\
The first hypothesis explores whether LLMs can effectively generate values for existing clinical features based on available patient data. To test this, we conducted an experiment where we compared LLM-generated values with those produced by traditional methods, specifically Gaussian-based permutation techniques. \textcolor{blue}{No noise was added to image data.} By comparing the outputs, we can assess whether LLMs are a reliable tool for generating patient-specific values that accurately reflect the clinical context.\\

\noindent
\textbf{Validating Hypothesis 2: Can LLMs Extend Their Knowledge to Generate New Features?}\\
Once we validate that LLMs can generate accurate values for existing features, we move on to the second hypothesis, which explores whether LLMs can extend their knowledge to create new clinically relevant features and generate their values. This requires a more sophisticated experiment design, which we carried out in the following stages:

\begin{itemize}
    \item \textbf{Stage 1: Analysing the Quality of Information Retrieval (see Section \ref{sec:experiment2}).} The first step in validating Hypothesis 2 involves understanding what data sources contribute to generating high-quality, contextually relevant information. To do this, we used data from two complementary sources: Wikipedia and Radiopaedia. 
    We tested which of these sources (Wikipedia, Radiopedia or both) give the LLM access to a more comprehensive clinical knowledge base, allowing it to generate better answers for both general and specialized queries (see Phase I in Figure \ref{fig:main}). \textcolor{blue}{Other sources, such as medical textbooks and clinical databases (e.g., UpToDate, PubMed, and professional guidelines), contain high-quality expert-reviewed content. However, their integration into our framework is limited by copyright and access restrictions, preventing their direct use in automated retrieval pipelines.}

    \item \textbf{Stage 2: Generating Prompts for New Feature Generation (see Section \ref{sec:experiment4}).} After understanding which data sources lead to better information retrieval, the next step was to design prompts that guide the LLM in generating new clinical features. Based on a set of questions given to radiologists, these prompts are tailored to retrieve information from Radiopedia and Wikipedia and to \textit{extract novel features from the LLM answers}. This extends the LLM’s understanding beyond existing features, allowing it to synthesize entirely new, contextually relevant features from the knowledge it has gained in Stage 1 (see Phase II in Figure \ref{fig:main}). \textcolor{blue}{In this stage, we designed retrieval-specific prompts that guide the LLM in generating context-aware features. Unlike approaches that fine-tune prompts through systematic training, we employed a trial-and-error method to iteratively refine the prompts based on observed model behavior.}

    \item \textbf{Stage 3: Populating the Generated Features with new Feature Values (see Section \ref{sec:experiment4}).} After identifying new features, we used the RAG framework to generate accurate values for each feature using the contextual knowledge established in previous phases. The RAG framework first retrieves relevant information from the Augmented Clinical Knowledge corpus, and then the generator model synthesizes this information to produce clinically coherent values. These generated values are then integrated into the original dataset, creating a richer, contextually enhanced dataset that aligns with the patient’s clinical profile and is ready for downstream analysis.

    \item \textbf{Stage 4: Evaluating the Effectiveness of the Newly Generated Features (see Section \ref{sec:experiment4})}. We carried out a comprehensive set of experiments to evaluate the new features generated by the LLMs. First, we \textbf{augmented the dataset with the newly generated features} and tested their impact on several machine learning models, including Decision Trees, Random Forest, XGBoost, and TabNet. The models were evaluated using metrics such as accuracy and AUC to determine the added value provided by the new features. We also performed \textbf{an ablation study} to further investigate the contribution of the Augmented Clinical Knowledge from Phase II. By isolating the ACK's influence on the LLM's ability to generate valuable features, we assessed its critical role in the feature-generation process. Finally, we conducted \textbf{a feature importance analysis} to identify which of the newly generated features had the greatest relevance to clinical outcomes. These findings were subsequently reviewed by medical experts to ensure the clinical validity and relevance of the generated features. \textcolor{blue}{In evaluating the effectiveness of the newly generated features, we initially report accuracy and AUC as general performance indicators. However, we recognize that in clinical applications, false negatives are particularly concerning, as missing relevant clinical information could impact diagnosis and treatment decisions. To address this, our results emphasize precision and recall, as they provide a more meaningful assessment of feature validity. Precision ensures that newly generated features are clinically relevant, while recall minimizes the risk of missing critical information. This dual focus allows us to balance overall model performance with medical reliability.}\\
\end{itemize}

\subsection{Parameters and Model Setup}

\textcolor{blue}{In our experiments, no fine-tuning was performed, as we leveraged a Retrieval-Augmented Generation approach that directly incorporates medical sources, eliminating the need for model-specific adaptation. Since each LLM has its own internal mechanisms for interpreting queries, there is no standardized method for prompt formulation. To ensure a fair evaluation, we used a consistent query formulation across all models without modifying LLM-specific prompt structures, preventing biases in prompt optimization and ensuring that each model’s response reflects its native ability to retrieve and synthesize clinical knowledge. Additionally, we set the LLM temperature to 0.1 to prioritize deterministic and clinically consistent outputs, as higher temperatures (e.g., 0.7 or above) occasionally led to clinically implausible outputs, such as symptoms unrelated to the patient’s condition. Given the sensitive nature of medical data augmentation, this choice was guided by preliminary tests, clinical requirements, and consistency considerations, ensuring that generated outputs align closely with retrieved domain-specific knowledge from Radiopaedia and the patient context.}

\textcolor{blue}{In the downstream evaluation of our augmented dataset, we employed default parameters for each machine learning model to ensure consistency and reproducibility. For Decision Trees and Random Forests, we used default settings, with Decision Trees having a maximum depth of 10 and Random Forests consisting of 100 trees, both utilizing the Gini impurity function for node splitting. XGBoost was configured with a learning rate of 0.01, a max depth of 6, and 100 estimators, using the binary:logistic setting to accommodate the binary nature of our classification tasks. For the TabNet model, we leveraged the PyTorch implementation from Dreamquark AI with its preset parameters, including an optimizer learning rate of 0.02, a batch size of 1024, and a decision dimension of 8 with 2 steps in the decision process.}




All experiments used high-performance computational resources, including NVIDIA 4090 24GB GPUs and Intel i9-13900K CPUs. These setups allowed for efficient training and testing of the machine learning models and rapid processing of the large datasets required for the experiments.

\subsection{Downstream Task Parameters}
\textcolor{blue}{
In the downstream evaluation of our augmented dataset, we employed default parameters for each machine learning model to ensure consistency and reproducibility. For Decision Trees and Random Forests, default settings were applied, Decision Trees having a maximum depth of 10 and Random Forests consisting of 100 trees, both using the Gini impurity function for node splitting. For XGBoost, default parameters included a learning rate of 0.01, a max depth of 6, and 100 estimators, with the model set to `binary:logistic` to accommodate the binary nature of our classification tasks. For the TabNet model, we used the PyTorch implementation from Dreamquark AI with its preset parameters. These include an optimiser learning rate of 0.02, a batch size of 1024, and a decision dimension of 8 with 2 steps in the decision process. 
}

\section{Results}

\subsection{Experiment I: Can LLMs Generate Contextual Values for Existing Features} \label{sec:generate-synthetic-values}

\begin{figure}
  \centering
  \includegraphics[width=\textwidth]{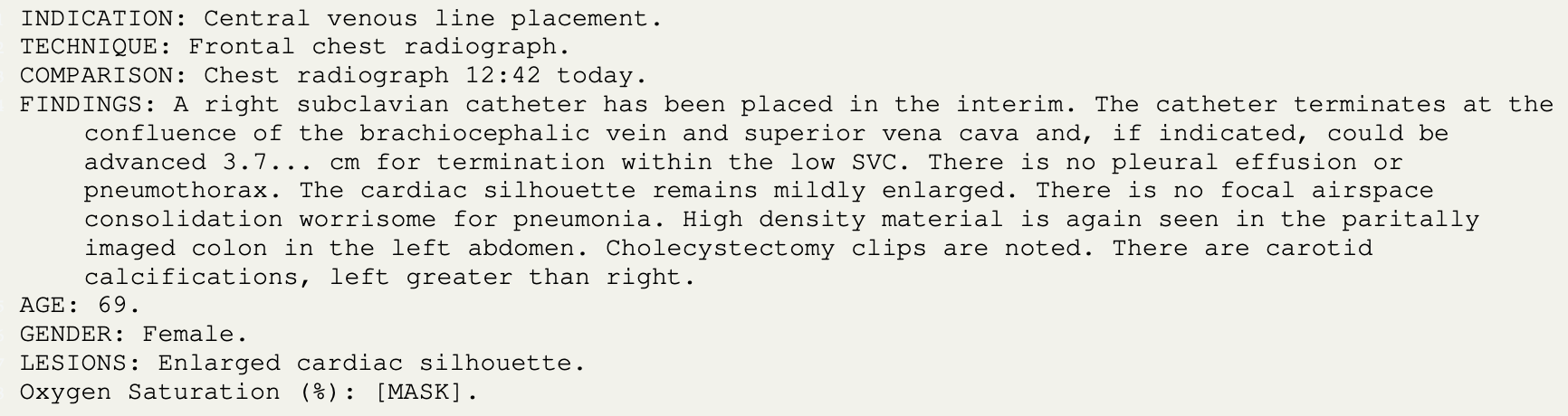}
  \caption{Example of a prompt used in Experiment I for generating feature values for Oxygen Saturation (\%).}
  \label{fig:exp1_specific}
\end{figure}

This experiment evaluated the potential of LLMs to accurately generate synthetic values for pre-existing clinical features within the MIMIC-IV dataset. The criticality of this task cannot be overstated, given the propensity of LLMs to produce ``hallucinations'' or inaccuracies that, within a medical context, could lead to significant ramifications for patient care. To mitigate these risks and ensure the generation of clinically relevant values, we employed a structured approach involving generating a prompt (Figure \ref{fig:exp1_specific}). We set the temperature parameter to $0.1$ and applied eight clinical features from the MIMIC-IV dataset (oxygen saturation, temperature, gender, age, heart rate, o2sat, systolic (sbp), and diastolic blood pressure (dbp)) to $799$ patient cases. 

We analyzed the capabilities of different types of LLMs to generate clinically relevant synthetic data. We tested nine different LLMs ranging from medical domain-specific models (such as BioGPT\citep{Luo2022}, ClinicalBERT\citep{Huang2019}, BioClinicalBERT\citep{alsentzer2019}, and Medtiron\citep{chen2023}) to more general models (Mistral\citep{jiang2023}, Zehyr\citep{tunstall2023}, Llama2\citep{touvron2023}, GPT-3.5, and GPT-4). Figure \ref{fig:exp1_specific} presents an example of a prompt to generate values for existing features based on the patient's clinical context.

To quantify the deviation of LLM-generated values from actual clinical data, we employed the Mean Squared Error (MSE) metric, comparing the generated values against the ground truth for each feature. We use this metric to indicate the LLM's accuracy and reliability in synthesizing clinical data.

We used a simple Gaussian distribution to generate clinical agnostic feature values as a baseline. 
\textcolor{blue}{To compare the performance of LLM-generated clinical feature values with a multimodal approach, we implemented a Multimodal Transformer (MMTF) inspired by ViT~\citep{dosovitskiy2020image}. The MMTF was trained using chest X-ray images and corresponding structured clinical features from the MIMIC-IV dataset. The model was optimized using a cross-entropy loss function and fine-tuned over 200 epochs with an Adam optimizer (learning rate = 0.0001). Evaluation was conducted using the same Mean Squared Error metric applied to LLM-generated values, ensuring a direct comparison between methods.}

The experiment was structured to run each query through the multimodal vision transformer, mirroring the approach taken with the LLMs to ensure comparability. Table~\ref{tab:aug-mse} reports the results using MIMIC-IV, where GPT-4 provides the best performance using mean MSE (i.e., average overall features).

\begin{table*}[!h]
\resizebox{\columnwidth}{!}{
\begin{tabular}{|l|c|c|c|c|c|c||c|}
\hline
Model                               & temperature & heartrate & resprate & o2sat & sbp      & dbp   & Mean \\ 
\hline
BioGPT\citep{Luo2022}                & 6.124     & 1.733     & 108.269  & 7.799 & 3.628    & 1.836 & 21.565 \\
ClinicalBERT\citep{Huang2019}        & 5.623     & 2.663     & 22.880   & 2.298 & 6.086    & 1.443 & 6.832  \\
BioClinicalBERT\citep{alsentzer2019} & 4.204     & 1.578     & 7.230    & {\bf 1.466}      & 5.007 & 1.661 & 3.524          \\
Mistral\citep{jiang2023}           & 3.635       & 1.920     & 5.842    & 7.824 & 2.284 & 3.056 & 4.093          \\
Zephyr\citep{tunstall2023}            & 2.144       & 1.541     & 3.060    & 4.146 & 1.770 & {\bf 1.700} & 2.393          \\
Llama2\citep{touvron2023}            & {\bf 1.318}       & 1.559     & 2.617    & 2.387 & 1.494 & 3.741 & 2.186          \\
Meditron\citep{chen2023}          & 2.420       & 2.057     & 4.858    & 6.359 & 2.205 & 4.071 & 3.662          \\
GPT-3.5       & 1.325       & {\bf 1.308}     & {\bf 1.159}    & 3.403 & 1.264 & 2.338 & 1.799          \\
GPT-4        & 1.966       & 1.344     & 1.298    & 1.722 & {\bf 1.251} & 1.705 & \textbf{1.548} \\ \hline
MMTF (supervised)  & 0.003       & 0.008     & 0.040    & 0.009 & 0.006 & 0.013 & 0.013          \\ \hline
Gaussian (baseline) & 2.074       & 2.085     & 2.027    & 1.987 & 2.092 & 1.934 & 2.033          \\ \hline
\end{tabular}
}
\caption{Mean Square Error for normalised clinical features.}
\label{tab:aug-mse}
\end{table*}

\subsection{Experiment II: Analysing the Quality of Information Retrieval}\label{sec:experiment2}

In Experiment II, we evaluated the effectiveness of information retrieval and data storage mechanisms by comparing LLMs' performance using data from Radiopaedia and Wikipedia. The experiment focused on two main aspects: the information sources and the data representation types.

Using the best LLM from Experiment I, GPT-4, we conducted queries using three configurations: Wikipedia alone, Radiopaedia alone, and combined. Due to the high costs of using GPT-4, we limited our experiments to a single query: "Atelectasis." The following conditions were tested: (a) all documents from the first result page of Wikipedia (approximately 10 documents); (b) all documents from the first result page of Radiopaedia (approximately 20 documents); (c) the top result from Wikipedia (1 document); (d) the top result from Radiopaedia (1 document); and (e) all documents from the first result pages of both Wikipedia and Radiopaedia (approximately 30 documents total). For each setting, we tested three types of data representation: NetworkX, Neo4j Vector databases, and Neo4j Graph databases. The goal was to retrieve information for the domain questions outlined in Phase II: (1) What is Atelectasis?, (2) What are the symptoms of Atelectasis?, (3) What can cause Atelectasis?, (4) What clinical features do radiologists need for diagnosing Atelectasis?, and (5) What are the radiographic features of Atelectasis?

\begin{table}[!htbp]
\renewcommand{\arraystretch}{1.2} 
\tiny 
\resizebox{\columnwidth}{!}{ 
\begin{tabular}{|l|p{0.8\textwidth}|}  
\hline
\multicolumn{1}{l}{\textbf{Question}} &
  \textbf{LLM Answer} \\ \hline
\multicolumn{2}{c}{\textbf{Networkx}} \\ \hline
\cellcolor{codegray}1 & \cellcolor{codegray}
  Atelectasis is a condition in which small areas of the \textbf{lung collapse} due to adjacent compression, passive atelectasis, or dependent atelectasis. It is usually seen on chest x-rays as small volume linear shadows and CT is often helpful to determine the cause or confirm that there is no proximal obstruction. \\
2 &
  Atelectasis is usually \textbf{asymptomatic}, but some people may experience \textbf{shortness of breath}, \textbf{chest pain}, and \textbf{coughing} \\
\cellcolor{codegray}3 & \cellcolor{codegray}
  Atelectasis can be caused by \textbf{adjacent compression}, \textbf{passive atelectasis}, and \textbf{dependent atelectasis} \\
4 &
  Radiologists need to look for small areas of \textbf{collapsed lung}, \textbf{small volume linear shadows}, and \textbf{adjacent compression} on chest x-rays to diagnose Atelectasis. CT scans may also be helpful to determine the cause or confirm that there is no proximal obstruction \\
\cellcolor{codegray}5 & \cellcolor{codegray}
  Atelectasis is usually seen on chest x-rays as \textbf{small volume linear shadows} \\ \hline
\multicolumn{2}{c}{\textbf{Neo4jVector}} \\ \hline
1 &
  Atelectasis is the \textbf{collapse or closure of a lung} resulting in \textbf{reduced or absent gas exchange}. It is usually unilateral, affecting part or all of one lung. It is a condition where the alveoli are deflated down to little or no volume. It can be caused by \textbf{normal exhalation} or various medical conditions. Atelectasis is not synonymous with a \textbf{pneumothorax}, although a \textbf{pneumothorax} can cause atelectasis. It can occur as a \textbf{post-operative} complication or as a result of \textbf{surfactant deficiency}. It is a common finding in chest X-rays and other radiological studies. The most common cause is \textbf{post-surgical} atelectasis, which develops in 75-90\% of people undergoing general \textbf{anesthesia} for a surgical procedure. Other causes include pulmonary \textbf{tuberculosis}, \textbf{obstruction of the airway}, \textbf{poor surfactant spreading}, and \textbf{compression of the lung} by a space-occupying lesion. Clinically significant atelectasis is visible on chest X-ray and may require further imaging or bronchoscopy for diagnosis. Treatment involves managing the underlying cause \\
\cellcolor{codegray}2 & \cellcolor{codegray}
  The symptoms of Atelectasis may include \textbf{cough} (not prominent), \textbf{chest pain} (not common), \textbf{breathing difficulty} (fast and shallow), \textbf{low oxygen saturation}, \textbf{pleural effusion} (transudate type), \textbf{cyanosis} (late sign), and \textbf{increased heart rate}. However, it is important to note that atelectasis can also be \textbf{asymptomatic} \\
3 &
  Atelectasis can be caused by various factors, including \textbf{post-surgical complications}, \textbf{surfactant deficiency}, \textbf{pulmonary tuberculosis}, \textbf{blockage of the airway}, \textbf{compression from external factors}, \textbf{poor surfactant spreading}, and certain risk factors such as type of \textbf{surgery}, \textbf{obesity}, \textbf{high oxygen}, and \textbf{presence of chronic obstructive pulmonary disease} or \textbf{asthma}. It can also occur due to passive atelectasis or adhesive atelectasis \\
\cellcolor{codegray}4 & \cellcolor{codegray}
  Radiologists need to look for \textbf{small volume linear shadows} on chest x-rays, usually peripherally or at the lung bases, to diagnose atelectasis. They can also use CT scans to accurately depict the involved lung and determine the cause. Additionally, the presence of the \textbf{crow feet sign}, which appears as linear bands radiating from a mass into adjacent lung tissue, can be a characteristic feature of round atelectasis. The clinical presentation of atelectasis can vary depending on the extent and speed of development, but common symptoms include \textbf{breathlessness}, \textbf{cough}, and \textbf{diminished breath sounds}. The underlying causes of atelectasis can include \textbf{adjacent compression}, passive atelectasis, and dependent atelectasis. Treatment involves managing the underlying cause and promoting deeper breathing if hypoventilation is the cause. \\
5 &
  Radiographic features of atelectasis include \textbf{small volume linear shadows} on chest x-rays, usually peripherally or at the lung bases. The underlying cause may also be visible. CT chest allows for a more accurate depiction of the involved lung and can help determine the cause. Round atelectasis can be seen on CT as linear bands radiating from a mass into adjacent lung tissue, resembling the \textbf{feet of a crow}. Atelectasis can also be classified based on underlying mechanism (resorptive, passive, compressive, cicatrisation, adhesive, gravity dependent) or morphology (linear, round, lung, lobar, segmental, subsegmental). The radiographic features vary depending on the type and mechanism of atelectasis. \\ \hline
\multicolumn{2}{c}{\textbf{Neo4jGraph}} \\ \hline
\cellcolor{codegray}1 & \cellcolor{codegray}
  Atelectasis is a condition where part or all of one lung collapses or closes, resulting in \textbf{reduced or absent gas exchange}. It can be caused by normal exhalation or various medical conditions. The alveoli, which are small air sacs in the lungs, become deflated down to little or no volume. Atelectasis is also known as a collapsed lung. \\
2 &
  \textcolor{red}{I'm sorry, but I don’t have the information to answer your question. It would be best to consult a medical professional for accurate information on the symptoms of Atelectasis} \\
\cellcolor{codegray}3 & \cellcolor{codegray}
  There are several factors that can cause atelectasis. Some common causes include \textbf{blockage of the airways} due to mucus, a \textbf{foreign object}, or a \textbf{tumor}, as well as lung diseases such as \textbf{chronic obstructive pulmonary disease (COPD)} or \textbf{pneumonia}. Additionally, certain medical procedures, such as \textbf{anesthesia} or \textbf{prolonged bed rest}, can also contribute to the development of atelectasis. It is important to consult with a healthcare professional for a proper diagnosis and treatment plan. \\
4 &
  To diagnose Atelectasis, radiologists typically look for certain clinical features. These may include signs such as \textbf{lung collapse}, \textbf{airway obstruction}, and \textbf{reduced lung volume}. Additionally, they may also consider the presence of symptoms such as \textbf{shortness of breath}, \textbf{coughing}, and \textbf{chest pain}. By evaluating these clinical features, radiologists can make an accurate diagnosis of Atelectasis \\
\cellcolor{codegray}5 & \cellcolor{codegray}
  \textcolor{red}{I'm sorry, but I don't have the information to answer your question about the radiographic features of Atelectasis.} \\ \hline
\multicolumn{1}{l}{} &
   \\ \hline
\end{tabular}
}
\caption{GPT-4 Responses Based on Wikipedia and Radiopaedia: Synthesis of Information Retrieved from the Top Search Results for Query "Atelectasis". 
\textcolor{blue}{The highlighted terms indicate areas where additional clinical context was retrieved to guide the feature generation process. While these terms alone do not directly translate into new features, they help DALL-M retrieve relevant medical knowledge, which is then used to synthesize new, context-aware features.}
}
\label{tab:exp2_topdocs_both}
\end{table}

\textcolor{blue}{The most effective retrieval configuration was determined by comparing different data sources (Wikipedia, Radiopaedia, or both) and storage methods (NetworkX, Neo4j Vector, and Neo4j Graph databases). The assessment was based on the completeness, accuracy, and contextual relevance of the retrieved information in answering domain-specific questions (e.g., symptoms, causes, and radiographic features of atelectasis). The best-performing setup, using both Wikipedia and Radiopaedia stored in a Neo4j Vector database, provided the most comprehensive and clinically meaningful responses.}

Our experiments show that the best data representation is the Neo4jVector (see Table \ref{tab:exp2_topdocs_both} for an example of LLM outputs). This is due to its ability to capture and retrieve semantic relationships within high-dimensional data efficiently. The Neo4j Vector database outperformed other methods by providing more contextually relevant and accurate responses to the domain questions. Its advanced indexing and search capabilities allowed for better integration of diverse information sources, ultimately enhancing the LLM's ability to generate high-quality, clinically relevant features. This demonstrates that leveraging a combination of broad and specialized knowledge stored in a semantically rich vector format significantly improves the performance of the LLM in clinical data augmentation tasks.

\subsection{Experiment III: Evaluating the Effectiveness of the Newly  Generated Features}\label{sec:experiment4}

Building on the initial assessment of LLMs to generate realistic values for existing clinical features, this experiment evaluates DALL-M's capability to create new, clinically relevant features to enhance the dataset comprehensively. This investigation involved processing $799$ patient case data points, originally comprising only eight features, to include new features and their corresponding synthetic values, as established in Experiment I. We termed this method \textit{Augmented}, containing 78 features.
Furthermore, we consulted medical experts to leverage clinical expertise and enrich the dataset. They reviewed the newly generated features and recommended including specific clinical features 
based on their professional experience. The subsequent generation of values, 
 denoted as \textit{Augmented with Expert Input}, yielded 91 features.

To rigorously assess the contribution of these augmented features to the dataset, three distinct analytical approaches were employed:
(1) We evaluated the augmented dataset using various ML models, including Decision Trees, Random Forest, XGBoost, and TabNet. This analysis used metrics such as accuracy and area under the curve (AUC);
(2) An ablation study was conducted to discern the impact of incorporating ACK into the prompt generation phase on the overall effectiveness of the augmentation process;
(3) The ML models employed underwent a feature importance analysis. Our medical experts then reviewed the results to validate the clinical relevance of the identified features.

\begin{table}[!h]
\centering
\small
\resizebox{\columnwidth}{!}{
\begin{tabular}{llccccccc}
\hline
\textbf{Feature set}                                                                                  & \textbf{Model} & \textbf{Accuracy}             & \textbf{AUC}                  & \textbf{Precision}            & \textbf{F-1}                  & \textbf{Recall}               & \multicolumn{1}{l}{\textbf{P-value}} & \textbf{\#Rel. F.} \\ \hline
                                                                                                      & DecisionTree   & 0.6676                        & 0.5666                        & 0.367                         & 0.3591                        & 0.352                         & 0.3021                               & 0                  \\
                                                                                                      & RandomForest   & 0.7484                        & 0.554                         & 0.6091                        & 0.2285                        & 0.1408                        & 0                                    & 0                  \\
                                                                                                      & XGBoost        & 0.7108                        & 0.5389                        & 0.3953                        & 0.2411                        & 0.1735                        & 0                                    & 8                  \\
\multirow{-4}{*}{\begin{tabular}[c]{@{}l@{}}Original \\ \# feat: 8\end{tabular}}                      & TabNet         & {\color[HTML]{000000} 0.6568} & {\color[HTML]{000000} 0.5250} & {\color[HTML]{000000} 0.3117} & {\color[HTML]{000000} 0.2743} & {\color[HTML]{000000} 0.2449} & 0.5000                               & 6                  \\ \hline
                                                                                                      & DecisionTree   & 0.7319                        & 0.6124                        & 0.4935                        & 0.4139                        & 0.3582                        & 0.0222                               & 6                  \\
                                                                                                      & RandomForest   & 0.7749                        & 0.6089                        & {\ul \textit{0.7067}}         & 0.3752                        & 0.2561                        & 0                                    & 3                  \\
                                                                                                      & XGBoost        & \textbf{0.7865}               & \textbf{0.6687}               & 0.6508                        & \textbf{0.5093}               & \textbf{0.4184}               & 0.0001                               & 49                 \\
\multirow{-4}{*}{\begin{tabular}[c]{@{}l@{}}Augmented\\ \# feat: 78\end{tabular}}                     & TabNet         & 0.7027                        & 0.5856                        & 0.4231                        & 0.375                         & 0.3367                        & 0.0293                               & 22                 \\ \hline
                                                                                                      & DecisionTree   & 0.7246                        & 0.6244                        & 0.4767                        & 0.4413                        & 0.4112                        & 0.2322                               & 5                  \\
                                                                                                      & RandomForest   & 0.7811                        & 0.6177                        & \textbf{0.7362}               & 0.3953                        & 0.2704                        & 0                                    & 4                  \\
                                                                                                      & XGBoost        & {\ul \textit{0.7838}}         & {\ul \textit{0.6636}}         & 0.6452                        & {\ul \textit{0.5000}}                  & 0.4082                        & 0.0002                               & {\ul \textit{56}}  \\
\multirow{-4}{*}{\begin{tabular}[c]{@{}l@{}}Augmented with\\ Expert input\\ \# feat: 91\end{tabular}} & TabNet         & 0.7243                        & 0.6232                        & 0.4762                        & 0.4396                        & 0.4082                        & 0.3974                               & 43                 \\ \hline
                                                                                                      & DecisionTree   & 0.7127                        & 0.6182                        & 0.4549                        & 0.4349                        & {\ul \textit{0.4173}}         & 0.6686                               & 1                  \\
                                                                                                      & RandomForest   & 0.7705                        & 0.6174                        & 0.6483                        & 0.4022                        & 0.2918                        & 0                                    & 1                  \\
                                                                                                      & XGBoost        & 0.7486                        & 0.6365                        & 0.5342                        & 0.4561                        & 0.398                         & 0.4082                               & 47                 \\
\multirow{-4}{*}{\begin{tabular}[c]{@{}l@{}}Ablation\\ No Prior Knowledge\\ \# feat: 91\end{tabular}} & TabNet         & 0.6541                        & 0.5754                        & 0.3636                        & 0.3846                        & 0.4082                        & 0.6799                               & \textbf{59}        \\ \hline
\end{tabular}
}
\caption{Performance of different feature sets on REFLACX dataset \citep{Lanfredi2021REFLACX} using CheXpert \citep{irvin2019chexpert} labels as groundtruth. The bold values represent the best-performing models, while the underlined values correspond to the second-best. (\#Rel. F. = Relevant Features)}
\label{tab:chexpert-perf}
\end{table}

\begin{figure*}[!h]
    \resizebox{\columnwidth}{!}{
    \includegraphics{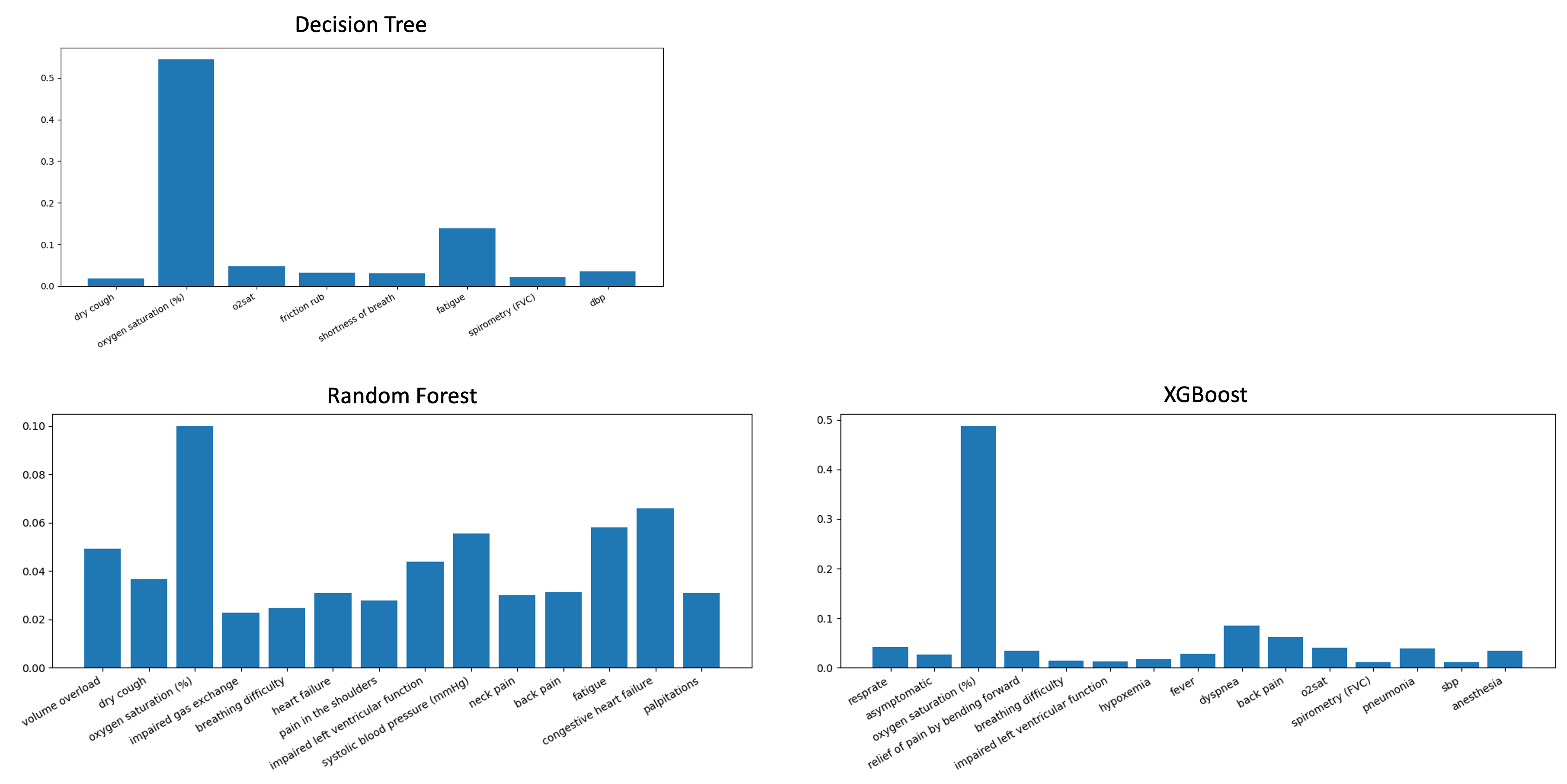}
    }
    \caption{Example of feature importance distribution for enlarged cardiac silhouette using Decision Trees, Random Forests and XGBoost.}
    \label{fig:dt}
\end{figure*}

\begin{figure*}[!h]
    \resizebox{\columnwidth}{!}{
    \includegraphics{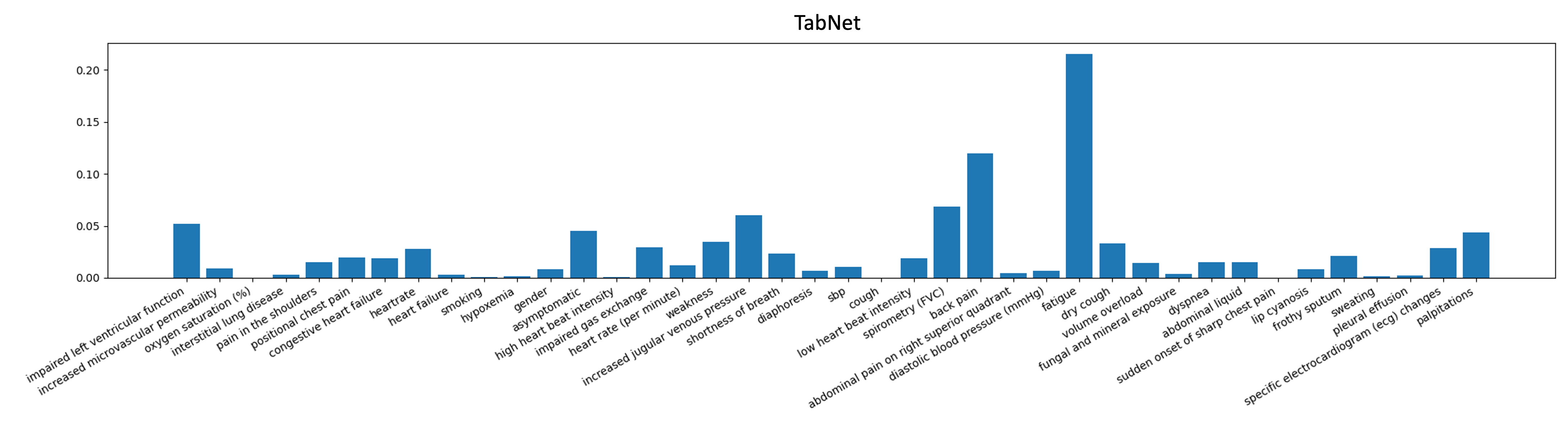}
    }
    \caption{Example of feature importance distribution for enlarged cardiac silhouette using TabNet}
    \label{fig:tabnet}
\end{figure*}

We documented the outcomes 
in Table~\ref{tab:chexpert-perf}. In the table, \textcolor{blue}{The "\#Rel. Features" column in Table 3 represents the number of features that were found to be important for each machine learning method. We measured feature importance across the generated and original features, and this column reflects how many features significantly contributed to the predictive performance of each algorithm.}
An example of feature importance distribution for Decision Trees, Random Forests, and XGBoost is presented in Figure \ref{fig:dt} and TabNet in Figure \ref{fig:tabnet}, with expert commentary on  the identified features' clinical validity discussed in  

\textcolor{blue}{DALL-M synthesized 70 new clinical features, later expanded to 91 after expert input. These features include lesion-specific attributes (e.g., pleural effusion severity, nodule shape descriptors), comorbidity indicators (e.g., history of chronic obstructive pulmonary disease), and additional symptom-related variables inferred from radiology reports.
In the Appendix, Table \ref{tab:all_features_augm} provides a detailed breakdown of these synthesized features.}

\begin{figure}
    \centering
    \includegraphics[width=1\linewidth]{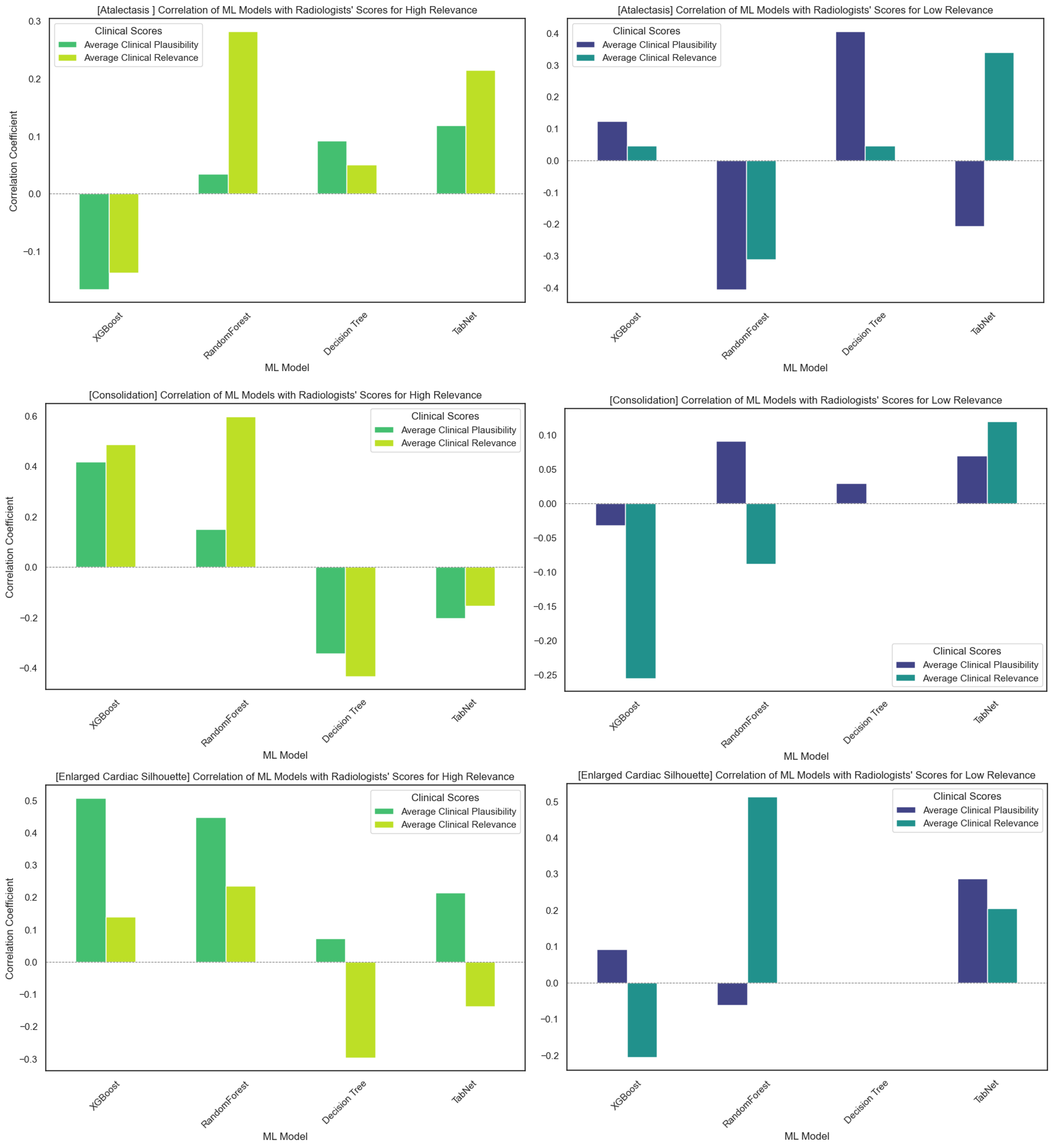}
    \caption{\textcolor{blue}{Correlation of Machine Learning Models with Radiologists' Clinical Scores for High and Low Relevance Features}}
    \label{fig:human_eval}
\end{figure}

\textcolor{blue}{
Figure \ref{fig:human_eval} presents the Pearson correlation coefficients between the feature importance scores of four machine learning models (XGBoost, Random Forest, Decision Tree, and TabNet) and the clinical evaluations provided by radiologists. Each panel corresponds to a specific condition—Atelectasis, Consolidation, and Enlarged Cardiac Silhouette—with separate plots for features rated as high relevance (left column) and low relevance (right column). The bars represent the average correlation with two clinical assessment criteria: clinical plausibility and clinical relevance for the set of features that were ranked least and highly important by two radiologists.}

\section{Discussion}
\label{sec:discussion}

Our experiments with the DALL-M framework highlight the accuracy of LLMs in generating synthetic clinical values and creating new clinically relevant features. The precision of models like GPT-4 and the enhancements from integrating expert insights are key focuses. We examine the impact of these advancements on predictive modeling, clinical decision-making, and patient care and contemplate the future role of AI in healthcare. The main contributions are as follows:\\

\noindent
\textbf{(1) LLMs Can Generate Clinically Relevant Synthetic Values}: Experiment~I demonstrates the potential of LLMs, particularly GPT-3.5 and GPT-4, to accurately generate synthetic values for existing clinical features. Using structured prompts, our evaluation through MSE suggests the models' ability to produce values close to real clinical data, with GPT-4 showing the best performance among LLMs.\\

\noindent
\textbf{(2) Multimodal Approaches Offer Further Improvements}: Integrating LLMs with other data modalities, such as images, could enhance the accuracy and relevance of generated clinical feature values. Although 
we explored this only 
in Experiment I, further research 
could incorporate these models in {\bf Phase III}. \\

\noindent
\textbf{(3) General LLMs perform better than Domain-Specific ones}: Experiment~I shows that general LLMs outperform domain-specific LLMs in generating synthetic clinical values. This may be due to several factors: (i) general LLMs are trained on a vast and diverse dataset, including ample medical literature, which may surpass the performance of models trained on narrower datasets, and (ii) general LLMs are updated regularly, leading to up-to-date information. \\

\noindent
\textbf{(4) DALL-M Improves Model Performance Despite the Complexity of Clinical Data:}
Experiment III demonstrates DALL-M’s ability to augment existing datasets with synthetic values and expand them with entirely new, clinically relevant features. Without medical expert input, DALL-M extended the eight original features to 78 features, significantly enhancing model performance.
\textcolor{blue}{Despite the inherent challenges of working with sparse and highly structured clinical tabular data, our results show a \textbf{25 percentage point increase in Precision and Recall} for XGBoost and a \textbf{16.5 percentage point improvement in F1-score}. These gains highlight the potential of DALL-M in addressing baseline performance limitations, particularly in real-world medical datasets that are often incomplete or imbalanced.} \textcolor{blue}{While Accuracy is a widely used metric, it can be misleading in medical datasets with class imbalances. In contrast, Precision, Recall, and F1-score are more clinically relevant, as they directly address the trade-off between false positives and false negatives. High Recall is crucial to avoid missing critical cases, while high Precision reduces unnecessary interventions, making these metrics more suitable for evaluating medical diagnostic tasks.} \textcolor{blue}{Additionally, data augmentation with expert input generated 13 additional features, totaling 91 features, leading to a further 3\% increase in Precision for Random Forests. These findings emphasize the practical benefits of context-aware feature augmentation in improving predictive modeling in healthcare. XGBoost consistently outperformed other models across most metrics, reaffirming the well-established effectiveness of tree-based models in handling tabular data~\citep{gorishniy2021revisiting,grinsztajn2022tree}.}

\noindent
\textbf{(5) Ablation studies confirm - Augmented Clinical Knowledge is vital:} The ablation study further confirms that DALL-M’s effectiveness is a result of its structured augmentation pipeline rather than reliance on expert-crafted prompts. \textcolor{blue}{While expert input refines the prompting process, its role is to enhance retrieval efficiency rather than inject external knowledge. The system's ability to generate high-quality synthetic values and new features primarily derives from its retrieval-augmented framework, which anchors responses in trusted medical sources.} This ensures that clinical consistency is achieved by design, independent of manual adjustments to prompts. As shown in Table~\ref{tab:chexpert-perf}, the absence of augmented clinical knowledge results in a notable performance decline across all models. This underscores the role of augmented clinical knowledge in enhancing the ability of LLMs to produce reliable and precise synthetic values. 
Nevertheless,
our approach requires ethical scrutiny for future iterations, to ensure accuracy, transparency, and mitigate biases \citep{Peng23,Thirunavukarasu23,Moreira2022fusion}.\\


\noindent
\textcolor{blue}{\textbf{(6) DALL-M Excels when Compared with Traditional Clinical Data Augmentation Methods}: Traditional augmentation techniques—such as SMOTE, GANs, and Gaussian noise injection—primarily manipulate existing data without introducing new, clinically meaningful information. SMOTE interpolates between existing samples, failing to enrich datasets with novel insights. GAN-based approaches generate synthetic data but often lack interpretability and can introduce artifacts. Gaussian perturbation methods inject randomness, risking implausible clinical relationships. DALL-M surpasses these methods by generating entirely new, context-aware features grounded in verified medical knowledge through RAG. This ensures medical consistency, interpretability, and enhanced predictive power. By integrating expert-driven contextualization with advanced AI techniques, DALL-M fundamentally redefines how clinical datasets are enriched, moving beyond simple augmentation to intelligent knowledge-driven feature expansion.}\\

\noindent
\textcolor{blue}{\textbf{(7) Alignment of Machine Learning Models with Radiologists' Clinical Assessments}: The correlation analysis indicates that Random Forest aligns most closely with radiologists' assessments of clinical relevance, particularly for high-relevance features in Consolidation and Atelectasis. This suggests that Random Forest effectively identifies diagnostically significant features, likely due to its ensemble approach, which enhances robustness and reduces overfitting. In contrast, TabNet tends to overestimate feature importance, especially for Enlarged Cardiac Silhouette, resulting in weaker correlations with radiologists' evaluations. This overestimation suggests that TabNet may capture broader patterns but struggles to differentiate between clinically critical features and less relevant ones, highlighting the importance of aligning machine learning feature selection with expert clinical judgment for improved interpretability and diagnostic support.
}

\section{\textcolor{blue}{Challenges, Limitations and Future Work}}

The DALL-M framework shows promise for enhancing clinical data, but faces several key challenges that need addressing.

\textcolor{blue}{While our results demonstrate the effectiveness of DALL-M, several challenges must be acknowledged for real-world deployment. First, LLM-generated data must maintain clinical consistency, as synthetic values may not always align with real-world patient distributions. Second, bias propagation remains a concern—LLMs trained on imbalanced medical datasets may reinforce existing disparities, leading to skewed predictions. Third, unlike real patient records, synthetic data lacks direct ground truth validation, making rigorous quality control essential. Additionally, the use of LLMs in clinical augmentation raises regulatory and ethical challenges, particularly regarding compliance with GDPR, and medical AI guidelines. Finally, LLMs may introduce hallucinations, where fabricated correlations or features could mislead predictive models if not carefully filtered. Future research should explore hybrid methods that combine LLM augmentation with domain expert validation} to mitigate these risks.

\textcolor{blue}{
The choice between specialized and general-purpose language models presents another dilemma. While GPT-4 currently outperforms medical-specific models, its effectiveness in specialized medical subfields remains uncertain. Additionally, the high computational costs of running advanced models like GPT-4 create practical barriers to widespread clinical adoption.
Ethical considerations also loom large. Though we use synthetic data to protect patient privacy, we must carefully monitor for potential biases and prevent misuse of generated clinical information. Moving forward, addressing these challenges while maintaining strict ethical guidelines will be crucial for successful clinical implementation.}

\textcolor{blue}{
For future work, we aim to explore bias mitigation strategies to ensure fairness and reliability in synthetic data generation. Techniques such as adversarial debiasing, reweighting methods, and fairness-aware augmentation will be investigated to address potential disparities introduced by LLM-generated clinical data.  Additionally, we plan to expand DALL-M's capabilities by integrating multimodal clinical data, including radiology images and electronic health records (EHRs). By leveraging multimodal fusion techniques, we can enhance the contextual depth of synthetic feature generation, improving its alignment with real-world medical scenarios.}

\section{Conclusion}

This study proposes two distinct strategies for augmenting clinical datasets. The first approach addresses the issue of missing values by generating realistic synthetic values using trained LLMs and multimodal transformers for supervised learning. This method ensures the dataset is complete and robust, allowing machine learning models to learn more effectively.

The second approach focuses on generating clinically relevant features by leveraging the combined power of LLMs and medical experts' domain knowledge. This method goes beyond simply filling in gaps; it creates new, valuable data points that enhance the overall quality of the dataset. By incorporating expert insights, we ensure the generated features are statistically sound and clinically meaningful.

Our results demonstrate that using the augmented dataset significantly improves model performance across various machine learning models. The features identified through our augmentation strategies are validated as clinically relevant, showing potential to enhance diagnostic accuracy and patient care. To the best of our knowledge, DALL-M is the only method capable of generating new context-aware features in tabular data. This capability is particularly impactful in fields where data is scarce, as it can significantly enhance the quality and quantity of available data, leading to better model performance and more informed decision-making.


\section{Source Code}

The DALL-M source code is publicly
available at: \\ \url{https://github.com/ChihchengHsieh/DALL-M}

\section{Acknowledgments}
The work reported in this article was partially supported by the Center of Data Science at Queensland University of Technology and was partially supported under the auspices of the UNESCO Chair on AI \& VR by national funds through Fundação para a Ciência e a Tecnologia with references DOI:10.54499/\-UIDB/\-50021/2020,
DOI:10.54499/DL57/2016/CP1368/\-CT0002
and 2022.09212.PTDC (XAVIER project).\\

\bibliographystyle{splncs04}

\begin{thebibliography}{10}
\providecommand{\url}[1]{\texttt{#1}}
\providecommand{\urlprefix}{URL }
\providecommand{\doi}[1]{https://doi.org/#1}

\bibitem{44}
Al-Bahrani, R., Jha, D., Kang, Q., Lee, S., Yang, Z., Liao, W.K., Agrawal, A., Choudhary, A.N.: Sigrnn: Synthetic minority instances generation in imbalanced datasets using a recurrent neural network. In: Proceedings of the International Conference on Pattern Recognition Applications and Methods. pp. 349--356 (2021)

\bibitem{alsentzer2019}
Alsentzer, E., Murphy, J.R., Boag, W., Weng, W.H., Jin, D., Naumann, T., McDermott, M.B.A.: Publicly available clinical bert embeddings (2019)

\bibitem{38}
Barak, B., Chaudhuri, K., Dwork, C., Kale, S., McSherry, F., Talwar, K.: Privacy, accuracy, and consistency too: a holistic solution to contingency table release. In: Proceedings of the twenty-sixth ACM SIGMOD-SIGACT-SIGART symposium on Principles of database systems. pp. 273--282 (2007)

\bibitem{30}
Bharadwaj, S., Gupta, P., Bhagwan, R., Guha, S.: Discovering related data at scale. Proceedings of the VLDB Endowment  \textbf{14}(8),  1392--1400 (2021)

\bibitem{bigolin2022reflacx}
Bigolin~Lanfredi, R., Zhang, M., Auffermann, W.F., Chan, J., Duong, P.A.T., Srikumar, V., Drew, T., Schroeder, J.D., Tasdizen, T.: Reflacx, a dataset of reports and eye-tracking data for localization of abnormalities in chest x-rays. Scientific data  \textbf{9}(1), ~350 (2022)

\bibitem{Lanfredi2021REFLACX}
Bigolin~Lanfredi, R., Zhang, M., Auffermann, W.F., Chan, J., Duong, P.A.T., Srikumar, V., Drew, T., Schroeder, J.D., Tasdizen, T.: Reflacx, a dataset of reports and eye-tracking data for localization of abnormalities in chest x-rays. Scientific Data  \textbf{9} (2022)

\bibitem{bisercic2023interpretable}
Bisercic, A., Nikolic, M., van~der Schaar, M., Delibasic, B., Lio, P., Petrovic, A.: Interpretable medical diagnostics with structured data extraction by large language models. arXiv preprint arXiv:2306.05052  (2023)

\bibitem{12}
Bogatu, A., Fernandes, A.A., Paton, N.W., Konstantinou, N.: Dataset discovery in data lakes. In: 2020 ieee 36th international conference on data engineering (icde). pp. 709--720. IEEE (2020)

\bibitem{borisov2022language}
Borisov, V., Se{\ss}ler, K., Leemann, T., Pawelczyk, M., Kasneci, G.: Language models are realistic tabular data generators. arXiv preprint arXiv:2210.06280  (2022)

\bibitem{20}
Cappuzzo, R., Papotti, P., Thirumuruganathan, S.: Creating embeddings of heterogeneous relational datasets for data integration tasks. In: Proceedings of the 2020 ACM SIGMOD international conference on management of data. pp. 1335--1349 (2020)

\bibitem{31}
Cappuzzo, R., Papotti, P., Thirumuruganathan, S.: Creating embeddings of heterogeneous relational datasets for data integration tasks. In: Proceedings of the 2020 ACM SIGMOD international conference on management of data. pp. 1335--1349 (2020)

\bibitem{42}
Chen, H., Jajodia, S., Liu, J., Park, N., Sokolov, V., Subrahmanian, V.: Faketables: Using gans to generate functional dependency preserving tables with bounded real data. In: Proceedings of the International Joint Conference on Artificial Intelligence. pp. 2074--2080 (2019)

\bibitem{chen2023huatuogpt}
Chen, J., Wang, X., Gao, A., Jiang, F., Chen, S., Zhang, H., Song, D., Xie, W., Kong, C., Li, J., et~al.: Huatuogpt-ii, one-stage training for medical adaption of llms. arXiv preprint arXiv:2311.09774  (2023)

\bibitem{19}
Chen, P., Sarkar, S., Lausen, L., Srinivasan, B., Zha, S., Huang, R., Karypis, G.: Hytrel: Hypergraph-enhanced tabular data representation learning. Advances in Neural Information Processing Systems  \textbf{36} (2024)

\bibitem{chen2021synthetic}
Chen, R.J., Lu, M.Y., Chen, T.Y., Williamson, D.F., Mahmood, F.: Synthetic data in machine learning for medicine and healthcare. Nature Biomedical Engineering  \textbf{5}(6),  493--497 (2021)

\bibitem{chen2023}
Chen, Z., Cano, A.H., Romanou, A., Bonnet, A., Matoba, K., Salvi, F., Pagliardini, M., Fan, S., Köpf, A., Mohtashami, A., Sallinen, A., Sakhaeirad, A., Swamy, V., Krawczuk, I., Bayazit, D., Marmet, A., Montariol, S., Hartley, M.A., Jaggi, M., Bosselut, A.: Meditron-70b: Scaling medical pretraining for large language models (2023)

\bibitem{25}
Chepurko, N., Marcus, R., Zgraggen, E., Fernandez, R.C., Kraska, T., Karger, D.: Arda: automatic relational data augmentation for machine learning. arXiv preprint arXiv:2003.09758  (2020)

\bibitem{Moreira2022fusion}
Chou, Y.L., Moreira, C., Bruza, P., Ouyang, C., Jorge, J.: Counterfactuals and causability in explainable artificial intelligence: Theory, algorithms, and applications. Information Fusion  \textbf{81},  59--83 (2022)

\bibitem{1}
Cui, L., Li, H., Chen, K., Shou, L., Chen, G.: Tabular data augmentation for machine learning: Progress and prospects of embracing generative ai. arXiv preprint arXiv:2407.21523  (2024)

\bibitem{26}
Dai, Y., El-Roby, A., Adeeb, E., Thaker, V.: {OmniMatch: Overcoming the Cold-Start Problem in Cross-Domain Recommendations using Auxiliary Reviews}. In: Proceedings of the 28th International Conference on Extending Database Technology (EDBT). pp. 80--91 (2025)

\bibitem{27}
Dong, Y., Takeoka, K., Xiao, C., Oyamada, M.: Efficient joinable table discovery in data lakes: A high-dimensional similarity-based approach. In: 2021 IEEE 37th International Conference on Data Engineering (ICDE). pp. 456--467. IEEE (2021)

\bibitem{28}
Dong, Y., Xiao, C., Nozawa, T., Enomoto, M., Oyamada, M.: Deepjoin: Joinable table discovery with pre-trained language models.(2022). URL: https://arxiv. org/abs/2212.07588. doi  \textbf{10} (2022)

\bibitem{dosovitskiy2020image}
Dosovitskiy, A., Beyer, L., Kolesnikov, A., Weissenborn, D., Zhai, X., Unterthiner, T., Dehghani, M., Minderer, M., Heigold, G., Gelly, S., et~al.: An image is worth 16x16 words: Transformers for image recognition at scale. arXiv preprint arXiv:2010.11929  (2020)

\bibitem{7}
Engelmann, J., Lessmann, S.: Conditional wasserstein gan-based oversampling of tabular data for imbalanced learning. Expert Systems with Applications  \textbf{174},  114582 (2021)

\bibitem{45}
Engelmann, J., Lessmann, S.: Conditional wasserstein gan-based oversampling of tabular data for imbalanced learning. Expert Systems with Applications  \textbf{174},  114582 (2021)

\bibitem{22}
Fan, G., Wang, J., Li, Y., Zhang, D., Miller, R.: Semantics-aware dataset discovery from data lakes with contextualized column-based representation learning. arXiv preprint arXiv:2210.01922  (2022)

\bibitem{2}
Fan, G., Wang, J., Li, Y., Zhang, D., Miller, R.J.: Semantics-aware dataset discovery from data lakes with contextualized column-based representation learning. Proceedings of the VLDB Endowment  \textbf{16}(7),  1726--1739 (2023)

\bibitem{3}
Fang, X., Xu, W., Tan, F., Zhang, J., Hu, Z., Qi, Y., Nickleach, S., Socolinsky, D., Sengamedu, S., Faloutsos, C.: Large language models (llms) on tabular data: Prediction, generation, and understanding—a survey. arxiv 2024. arXiv preprint arXiv:2402.17944  (2024)

\bibitem{33}
Glass, M., Wu, X., Naik, A.R., Rossiello, G., Gliozzo, A.: Retrieval-based transformer for table augmentation. arXiv preprint arXiv:2306.11843  (2023)

\bibitem{gonzales2023synthetic}
Gonzales, A., Guruswamy, G., Smith, S.R.: Synthetic data in health care: A narrative review. PLOS Digital Health  \textbf{2}(1),  e0000082 (2023)

\bibitem{gorishniy2021revisiting}
Gorishniy, Y., Rubachev, I., Khrulkov, V., Babenko, A.: Revisiting deep learning models for tabular data. Advances in Neural Information Processing Systems  \textbf{34},  18932--18943 (2021)

\bibitem{grinsztajn2022tree}
Grinsztajn, L., Oyallon, E., Varoquaux, G.: Why do tree-based models still outperform deep learning on typical tabular data?  \textbf{35},  507--520 (2022)

\bibitem{gu2021domain}
Gu, Y., Tinn, R., Cheng, H., Lucas, M., Usuyama, N., Liu, X., Naumann, T., Gao, J., Poon, H.: Domain-specific language model pretraining for biomedical natural language processing. ACM Transactions on Computing for Healthcare (HEALTH)  \textbf{3}(1),  1--23 (2021)

\bibitem{han2023medalpaca}
Han, T., Adams, L.C., Papaioannou, J.M., Grundmann, P., Oberhauser, T., L{\"o}ser, A., Truhn, D., Bressem, K.K.: Medalpaca--an open-source collection of medical conversational ai models and training data. arXiv preprint arXiv:2304.08247  (2023)

\bibitem{21}
Hu, X., Wang, S., Qin, X., Lei, C., Shen, Z., Faloutsos, C., Katsifodimos, A., Karypis, G., Wen, L., Philip, S.Y.: Automatic table union search with tabular representation learning. In: Findings of the Association for Computational Linguistics: ACL 2023. pp. 3786--3800 (2023)

\bibitem{Huang2019}
Huang, K., Altosaar, J., Ranganath, R.: Clinicalbert: Modeling clinical notes and predicting hospital readmission (2019). \doi{10.48550/ARXIV.1904.05342}, \url{https://arxiv.org/abs/1904.05342}

\bibitem{touvron2023}
Hugo~Touvron, e.a.: Llama 2: Open foundation and fine-tuned chat models (2023)

\bibitem{irvin2019chexpert}
Irvin, J., Rajpurkar, P., Ko, M., Yu, Y., Ciurea-Ilcus, S., Chute, C., Marklund, H., Haghgoo, B., Ball, R., Shpanskaya, K., et~al.: Chexpert: A large chest radiograph dataset with uncertainty labels and expert comparison. In: Proceedings of the AAAI conference on artificial intelligence. vol.~33, pp. 590--597 (2019)

\bibitem{jiang2023}
Jiang, A.Q., Sablayrolles, A., Mensch, A., Bamford, C., Chaplot, D.S., de~las Casas, D., Bressand, F., Lengyel, G., Lample, G., Saulnier, L., Lavaud, L.R., Lachaux, M.A., Stock, P., Scao, T.L., Lavril, T., Wang, T., Lacroix, T., Sayed, W.E.: Mistral 7b (2023)

\bibitem{Johnson2021MIMIC_IV}
Johnson, A., Bulgarelli, L., Pollard, T., Horng, S., Celi, L.A., Mark, R.: Mimic-iv  (2021)

\bibitem{41}
Jordon, J., Yoon, J., Van Der~Schaar, M.: Pate-gan: Generating synthetic data with differential privacy guarantees. In: International conference on learning representations (2018)

\bibitem{49}
Kanter, J.M., Veeramachaneni, K.: Deep feature synthesis: Towards automating data science endeavors. In: 2015 IEEE international conference on data science and advanced analytics (DSAA). pp. 1--10. IEEE (2015)

\bibitem{50}
Katz, G., Shin, E.C.R., Song, D.: Explorekit: Automatic feature generation and selection. In: 2016 IEEE 16th international conference on data mining (ICDM). pp. 979--984. IEEE (2016)

\bibitem{17}
Khatiwada, A., Fan, G., Shraga, R., Chen, Z., Gatterbauer, W., Miller, R.J., Riedewald, M.: Santos: Relationship-based semantic table union search. Proceedings of the ACM on Management of Data  \textbf{1}(1),  1--25 (2023)

\bibitem{36}
Khatiwada, A., Shraga, R., Gatterbauer, W., Miller, R.J.: Integrating data lake tables. Proceedings of the VLDB Endowment  \textbf{16}(4),  932--945 (2022)

\bibitem{46}
Kim, J., Lee, C., Shin, Y., Park, S., Kim, M., Park, N., Cho, J.: Sos: Score-based oversampling for tabular data. In: Proceedings of the 28th ACM SIGKDD Conference on Knowledge Discovery and Data Mining. pp. 762--772 (2022)

\bibitem{li2024llava}
Li, C., Wong, C., Zhang, S., Usuyama, N., Liu, H., Yang, J., Naumann, T., Poon, H., Gao, J.: Llava-med: Training a large language-and-vision assistant for biomedicine in one day. Advances in Neural Information Processing Systems  \textbf{36} (2024)

\bibitem{4}
Li, P., He, Y., Yashar, D., Cui, W., Ge, S., Zhang, H., Rifinski~Fainman, D., Zhang, D., Chaudhuri, S.: Table-gpt: Table fine-tuned gpt for diverse table tasks. Proceedings of the ACM on Management of Data  \textbf{2}(3),  1--28 (2024)

\bibitem{29}
Liang, J., Lei, C., Qin, X., Zhang, J., Katsifodimos, A., Faloutsos, C., Rangwala, H.: Featnavigator: Automatic feature augmentation on tabular data. arXiv preprint arXiv:2406.09534  (2024)

\bibitem{14}
Limaye, G., Sarawagi, S., Chakrabarti, S.: Annotating and searching web tables using entities, types and relationships. Proceedings of the VLDB Endowment  \textbf{3}(1-2),  1338--1347 (2010)

\bibitem{48}
Lin, Y., Ding, B., Jagadish, H., Zhou, J.: Smartfeat: Efficient feature construction through feature-level foundation model interactions. arXiv preprint arXiv:2309.07856  (2023)

\bibitem{liu2024survey}
Liu, L., Yang, X., Lei, J., Liu, X., Shen, Y., Zhang, Z., Wei, P., Gu, J., Chu, Z., Qin, Z., et~al.: A survey on medical large language models: Technology, application, trustworthiness, and future directions. arXiv preprint arXiv:2406.03712  (2024)

\bibitem{8}
Liu, T., Fan, J., Tang, N., Li, G., Du, X.: Controllable tabular data synthesis using diffusion models. Proceedings of the ACM on Management of Data  \textbf{2}(1),  1--29 (2024)

\bibitem{lu2022clinicalt5}
Lu, Q., Dou, D., Nguyen, T.: Clinicalt5: A generative language model for clinical text. In: Findings of the Association for Computational Linguistics: EMNLP 2022. pp. 5436--5443 (2022)

\bibitem{Luo2022}
Luo, R., Sun, L., Xia, Y., Qin, T., Zhang, S., Poon, H., Liu, T.Y.: Biogpt: generative pre-trained transformer for biomedical text generation and mining. Briefings in Bioinformatics  \textbf{23}(6) (Sep 2022). \doi{10.1093/bib/bbac409}, \url{http://dx.doi.org/10.1093/bib/bbac409}

\bibitem{Xu2023}
{M. Xu}: {Medicalgpt: Training medical gpt model}. \url{ https://github.com/shibing624/MedicalGPT}, 2023

\bibitem{37}
Miller, R.J.: Open data integration. Proceedings of the VLDB Endowment  \textbf{11}(12),  2130--2139 (2018)

\bibitem{11}
Nargesian, F., Zhu, E., Pu, K.Q., Miller, R.J.: Table union search on open data. Proceedings of the VLDB Endowment  \textbf{11}(7),  813--825 (2018)

\bibitem{6}
Nazabal, A., Olmos, P.M., Ghahramani, Z., Valera, I.: Handling incomplete heterogeneous data using vaes. Pattern Recognition  \textbf{107},  107501 (2020)

\bibitem{uk_biobank}
Ollier, W., Sprosen, T., Peakman, T.: Uk biobank: from concept to reality. Pharmacogenomics  \textbf{6}(6),  639--646 (2005)

\bibitem{40}
Park, N., Mohammadi, M., Gorde, K., Jajodia, S., Park, H., Kim, Y.: Data synthesis based on generative adversarial networks. Proceedings of the VLDB Endowment  \textbf{11}(10),  1071--1083 (2018)

\bibitem{Peng23}
Peng, C., Yang, X., Chen, A., et~al.: A study of generative large language model for medical research and healthcare. npj Digital Medicine  \textbf{6}(1), ~210 (2023)

\bibitem{radiopedia}
Radiopedia:  (2023), \url{https://radiopaedia.org/}

\bibitem{13}
Sarma, A.D., Fang, L., Gupta, N., Halevy, A.Y., Lee, H., Wu, F., Xin, R., Yu, C.: Finding related tables. In: SIGMOD Conference. vol.~10, pp. 2213836--2213962 (2012)

\bibitem{singhal2023large}
Singhal, K., Azizi, S., Tu, T., Mahdavi, S.S., Wei, J., Chung, H.W., Scales, N., Tanwani, A., Cole-Lewis, H., Pfohl, S., et~al.: Large language models encode clinical knowledge. Nature  \textbf{620}(7972),  172--180 (2023)

\bibitem{54}
Stekhoven, D.J., B{\"u}hlmann, P.: Missforest—non-parametric missing value imputation for mixed-type data. Bioinformatics  \textbf{28}(1),  112--118 (2012)

\bibitem{5}
Tang, X., Zong, Y., Phang, J., Zhao, Y., Zhou, W., Cohan, A., Gerstein, M.: Struc-bench: Are large language models really good at generating complex structured data? arXiv preprint arXiv:2309.08963  (2023)

\bibitem{theodorou2023synthesize}
Theodorou, B., Xiao, C., Sun, J.: Synthesize high-dimensional longitudinal electronic health records via hierarchical autoregressive language model. Nature communications  \textbf{14}(1), ~5305 (2023)

\bibitem{Thirunavukarasu23}
Thirunavukarasu, A.J., Ting, D.S.J., Elangovan, K., Gutierrez, L., Tan, T.F., Ting, D.S.W.: Large language models in medicine. Nature Medicine  \textbf{29}(8),  1930--1940 (2023)

\bibitem{tian2023chimed}
Tian, Y., Gan, R., Song, Y., Zhang, J., Zhang, Y.: Chimed-gpt: A chinese medical large language model with full training regime and better alignment to human preferences. arXiv preprint arXiv:2311.06025  (2023)

\bibitem{tunstall2023}
Tunstall, L., Beeching, E., Lambert, N., Rajani, N., Rasul, K., Belkada, Y., Huang, S., von Werra, L., Fourrier, C., Habib, N., Sarrazin, N., Sanseviero, O., Rush, A.M., Wolf, T.: Zephyr: Direct distillation of lm alignment (2023)

\bibitem{53}
Van~Buuren, S., Groothuis-Oudshoorn, K.: Mice: Multivariate imputation by chained equations in r. Journal of statistical software  \textbf{45},  1--67 (2011)

\bibitem{51}
Wang, D., Xiao, M., Wu, M., Wang, P., Zhou, Y., Fu, Y.: Reinforcement-enhanced autoregressive feature transformation: Gradient-steered search in continuous space for postfix expressions. In: Advances in Neural Information Processing Systems. vol.~36 (2023)

\bibitem{wang2023clinicalgpt}
Wang, G., Yang, G., Du, Z., Fan, L., Li, X.: Clinicalgpt: large language models finetuned with diverse medical data and comprehensive evaluation. arXiv preprint arXiv:2306.09968  (2023)

\bibitem{55}
Wang, Z., Sun, J.: Transtab: Learning transferable tabular transformers across tables. In: Advances in Neural Information Processing Systems. vol.~35, pp. 2902--2915 (2022)

\bibitem{wikipedia}
Wikipedia:  (2023), \url{https://wikipedia.org/}

\bibitem{52}
Xiao, M., Wang, D., Wu, M., Wang, P., Zhou, Y., Fu, Y.: Beyond discrete selection: Continuous embedding space optimization for generative feature selection. In: 2023 IEEE International Conference on Data Mining (ICDM). pp. 688--697 (2023)

\bibitem{47}
Xu, L., Skoularidou, M., Cuesta-Infante, A., Veeramachaneni, K.: Modeling tabular data using conditional gan. In: Advances in Neural Information Processing Systems (NeurIPS). vol.~32 (2019)

\bibitem{9}
Yakout, M., Ganjam, K., Chakrabarti, K., Chaudhuri, S.: Infogather: entity augmentation and attribute discovery by holistic matching with web tables. In: Proceedings of the 2012 ACM SIGMOD International Conference on Management of Data. pp. 97--108 (2012)

\bibitem{18}
Yakout, M., Ganjam, K., Chakrabarti, K., Chaudhuri, S.: Infogather: entity augmentation and attribute discovery by holistic matching with web tables. In: Proceedings of the 2012 ACM SIGMOD International Conference on Management of Data. pp. 97--108 (2012)

\bibitem{yang2022large}
Yang, X., Chen, A., PourNejatian, N., Shin, H.C., Smith, K.E., Parisien, C., Compas, C., Martin, C., Costa, A.B., Flores, M.G., et~al.: A large language model for electronic health records. NPJ digital medicine  \textbf{5}(1), ~194 (2022)

\bibitem{39}
Zhang, J., Cormode, G., Procopiuc, C.M., Srivastava, D., Xiao, X.: Privbayes: Private data release via bayesian networks. ACM Transactions on Database Systems (TODS)  \textbf{42}(4),  1--41 (2017)

\bibitem{16}
Zhang, L., Zhang, S., Balog, K.: {Table2Vec: Neural Word and Entity Embeddings for Table Population and Retrieval}. In: Proceedings of the 42nd International ACM SIGIR Conference on Research and Development in Information Retrieval (2019)

\bibitem{15}
Zhang, S., Balog, K.: {Entitables: Smart Assistance for Entity-Focused Tables}. In: Proceedings of the 40th International ACM SIGIR Conference on Research and Development in Information Retrieval (2017)

\bibitem{10}
Zhang, S., Balog, K.: Web table extraction, retrieval and augmentation. In: Proceedings of the 42nd International ACM SIGIR Conference on Research and Development in Information Retrieval. pp. 1409--1410 (2019)

\bibitem{43}
Zhang, Y., Zaidi, N., Zhou, J., Li, G.: Interpretable tabular data generation. Knowledge and Information Systems  \textbf{65}(7),  2935--2963 (2023)

\bibitem{32}
Zhao, Z., Castro~Fernandez, R.: Leva: Boosting machine learning performance with relational embedding data augmentation. In: Proceedings of the 2022 International Conference on Management of Data. pp. 1504--1517 (2022)

\bibitem{23}
Zhu, E., Deng, D., Nargesian, F., Miller, R.J.: Josie: Overlap set similarity search for finding joinable tables in data lakes. In: Proceedings of the 2019 International Conference on Management of Data. pp. 847--864 (2019)

\bibitem{24}
Zhu, E., Nargesian, F., Pu, K.Q., Miller, R.J.: Lsh ensemble: internet-scale domain search. Proceedings of the VLDB Endowment  \textbf{9}(12),  1185--1196 (2016)

\bibitem{zhu2023chatmed}
Zhu, W., Wang, X.: Chatmed: A chinese medical large language model (2023)

\end{thebibliography}

\appendix
\section{Prompt for Synthetic Clinical Feature Value Generation}\label{app1}

\begin{figure}[!h]
  \centering
  \includegraphics[width=\textwidth]{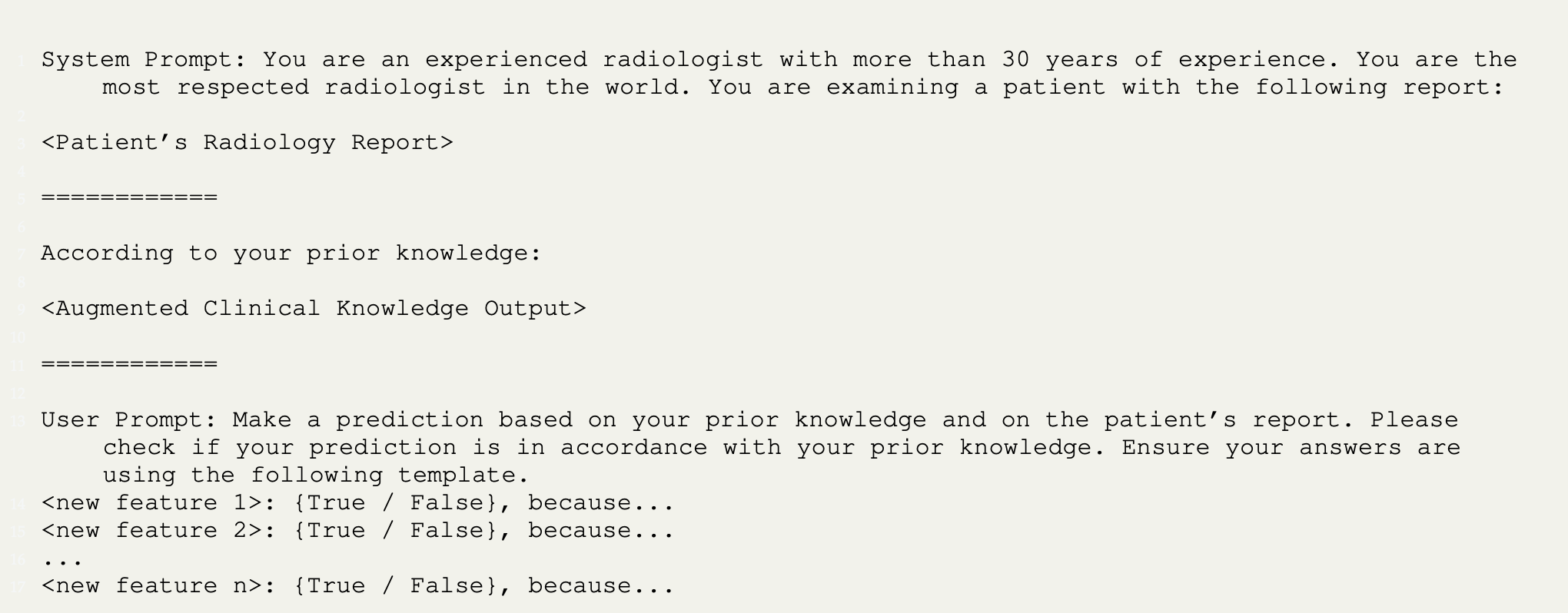}
  \caption{General template of the prompt used in the Synthetic Clinical Feature Value Generation Module with $n$ new features generated by the Augmented Clinical Knowledge Module. For a concrete example, please refer to Appendix Figure \ref{fig:synthetic_generation}.}
  \label{fig:exp3_general}
\end{figure}

Figure~\ref{fig:synthetic_generation} illustrates a concrete example of the final prompt used in the Synthetic Clinical Feature Value Generation Module. This prompt leverages the patient’s clinical report, extracted contextual information, and the expert knowledge synthesized in the Augmented Clinical KnowledgeModule to generate clinically relevant synthetic values for various features. The LLM is instructed to take on the role of an experienced radiologist, using its domain-specific knowledge to make predictions for specific clinical features such as “breathing difficulty” or “pleural effusion.”

The prompt begins by providing the patient's detailed report, including radiological findings and clinical impressions. Next, it outlines the relevant prior knowledge regarding the condition (in this case, atelectasis), covering its symptoms, potential causes, and diagnostic criteria. The LLM is then asked to assess whether the patient’s clinical presentation aligns with known patterns of atelectasis and other relevant conditions.

The final section of the prompt structures the LLM’s responses by using a template with binary feature values (e.g., True/False) accompanied by a rationale. This ensures that the generated values are not only accurate but also explainable, with each prediction supported by clinical reasoning. For example, the LLM might indicate “breathing difficulty: True, because the patient's report mentions fast and shallow breathing, which is a typical symptom of atelectasis.”

This structured and context-aware prompt enables the LLM to generate patient-specific values for existing features and introduce new clinically relevant features, enhancing the overall quality of the dataset for downstream analyses and machine learning applications.

\begin{figure*}[t]
\begin{codeframe}
\begin{lstlisting}[language=]
You are an experienced radiologist with more than 30 years of experience. You are the most respected radiologist in the world. You are examining a patient with the following report:

EXAMINATION:  PA AND LATERAL CHEST RADIOGRAPHS  INDICATION:  yearold male with weakness. Evaluate for pneumonia.  TECHNIQUE:  PA and lateral chest radiographs  COMPARISON:  Multiple prior chest radiographs  FINDINGS:   Compared with the previous examination there is increased diffuse interstitial opacities with a linear consolidation in the right lower lung compatible with atelectasis. There is also a focal opacity in the left for lower lung and retrocardiac region with associated small pleural effusion better seen in the lateral view.  Calcified granulomas are redemonstrated more prominently in the left apex. Moderate cardiomegaly is stable. Rightward deviation of the trachea is also unchanged.  There is no pneumothorax.  IMPRESSION:   Patchy opacities in the left lower lung and retrocardiac region may represent left lower lobe pneumonia on the background of interstitial pulmonary edema. LESIONS: Enlarged cardiac silhouette Atelectasis. AGE: 93. GENDER: Male. 
=========
According to your prior knowledge:

The symptoms associated with atelectasis may include cough (not prominent), chest pain (not common), breathing difficulty (fast and shallow), low oxygen saturation, pleural effusion (transudate type), cyanosis (late sign), and increased heart rate. However, atelectasis can also be asymptomatic. Fever is not a symptom of atelectasis.
Atelectasis can be caused by various medical conditions, including post-surgical complications, surfactant deficiency, and poor surfactant spreading during inspiration. It can also be caused by blockage of a bronchiole or bronchus, such as by a foreign body, mucus plug, tumor, or compression from the outside. Risk factors for atelectasis include certain types of surgery, muscle relaxation, obesity, high oxygen, lower lung segments, age, chronic obstructive pulmonary disease (COPD), asthma, and type of anesthetic.
The relevant symptoms for atelectasis include cough (not prominent), chest pain (not common), breathing difficulty (fast and shallow), low oxygen saturation, pleural effusion (transudate type), cyanosis (late sign), and increased heart rate.
The relevant clinical signs for the etiological diagnosis of atelectasis may include cough, chest pain (not common), breathing difficulty (fast and shallow), low oxygen saturation, pleural effusion (transudate type), cyanosis (late sign), and increased heart rate. However, it is important to note that atelectasis may also be asymptomatic.
The relevant laboratory data for the etiological diagnosis of atelectasis are not provided in the given information.
The relevant clinical characteristics for the etiological diagnosis of atelectasis include cough (not prominent), chest pain (not common), breathing difficulty (fast and shallow), low oxygen saturation, pleural effusion (transudate type), cyanosis (late sign), and increased heart rate. It is important to note that atelectasis does not cause fever. The underlying causes of atelectasis can include adjacent compression, passive atelectasis, dependent atelectasis, and poor surfactant spreading. Risk factors for atelectasis include type of surgery, use of muscle relaxation, obesity, high oxygen, lower lung segments, age, presence of chronic obstructive pulmonary disease or asthma, and type of anesthetic. Diagnosis of atelectasis is generally confirmed through chest X-ray, which may show lung opacification and/or loss of lung volume. Additional imaging modalities such as CT chest or bronchoscopy may be necessary to determine the cause of atelectasis.
The patient's personal relevant history for the etiological diagnosis of atelectasis includes post-surgical atelectasis as a common cause, as well as pulmonary tuberculosis, smoking, and old age as risk factors. Other factors associated with the development of atelectasis include the presence of chronic obstructive pulmonary disease or asthma, and the type of anesthesia used. The diagnosis of atelectasis is generally confirmed through chest X-ray, which shows small volume linear shadows, usually peripherally or at the lung bases. CT chest or bronchoscopy may be necessary to determine the cause or confirm the absence of proximal obstruction.\\
=========
Make a prediction based on your prior knowledge and on the patient's report. Please check if your prediction is in accordance with your prior knowledge. 
 Ensure your answers are using following template.
1. anesthesia: {True / False}, because...
2. asthma: {True / False}, because...\\
3. asymptomatic: {True / False}, because...
4. breathing difficulty: {True / False}, because...
5. chest pain: {True\textbackslash False}, because...
6. chronic obstructive pulmonary disease: {True /  False}, because...
7. cough: {True / False}, because...
8. cyanosis: {True / False}, because...
9. fever: {True / False}, because...
10. pleural effusion: {True / False}, because...
11. pulmonary tuberculosis: {True / False}, because...
12. small volume linear shadows: {True / False}, because...
13. smoking: {True\textbackslash False}, because...
14. oxygen saturation: {True\textbackslash False}, because...
...
\end{lstlisting}
\end{codeframe}
   \caption{Concrete example of the prompt used in the Synthetic Clinical Feature Value Generation Module after the Augmented Clinical Knowledge Module generated new features}\label{fig:synthetic_generation}
\end{figure*}

\section{\textcolor{blue}{Augmented Feature List}}

\textcolor{blue}{A comprehensive list of the 70 new features generated by DALL-M is provided, including their clinical category, average relevance score, and novelty assessment as determined by radiologists.}

\begin{table}[]
\resizebox{\columnwidth}{!}{
\begin{tabular}{|lll|}
\hline
\multicolumn{3}{|c|}{\textbf{Augmented Features}}                                                                                                    \\ \hline
\multicolumn{1}{|l|}{abdominal liquid}                          & \multicolumn{1}{l|}{fungal and mineral exposure} & pleural thickening               \\ \hline
\multicolumn{1}{|l|}{abdominal pain on right superior quadrant} & \multicolumn{1}{l|}{gender}                      & pneumonia                        \\ \hline
\multicolumn{1}{|l|}{age}                                       & \multicolumn{1}{l|}{heart failure}               & pneumothorax                     \\ \hline
\multicolumn{1}{|l|}{anesthesia}                                & \multicolumn{1}{l|}{heart rate}                  & positional chest pain            \\ \hline
\multicolumn{1}{|l|}{asymptomatic}                              & \multicolumn{1}{l|}{hiatal hernia}               & pulmonary tuberculosis           \\ \hline
\multicolumn{1}{|l|}{back pain}                                 & \multicolumn{1}{l|}{high bone density}           & pulsus paradoxus                 \\ \hline
\multicolumn{1}{|l|}{bone fractures}                        & \multicolumn{1}{l|}{high heart beat intensity}            & radiation to the trapezius ridge   \\ \hline
\multicolumn{1}{|l|}{breathing difficulty}                  & \multicolumn{1}{l|}{hypertensive crisis}                  & relief of pain by bending forward  \\ \hline
\multicolumn{1}{|l|}{cardiac tamponade}                         & \multicolumn{1}{l|}{hypoxemia}                   & relief of pain by inspiration    \\ \hline
\multicolumn{1}{|l|}{chest Pain}                                & \multicolumn{1}{l|}{impaired gas exchange}       & respiratory failure              \\ \hline
\multicolumn{1}{|l|}{chronic obstructive pulmonary disease} & \multicolumn{1}{l|}{impaired left ventricular function}   & respiratory rate                   \\ \hline
\multicolumn{1}{|l|}{congestive heart failure}              & \multicolumn{1}{l|}{increased jugular venous pressure}    & shortness of breath                \\ \hline
\multicolumn{1}{|l|}{cough}                                 & \multicolumn{1}{l|}{increased microvascular permeability} & small volume linear shadows        \\ \hline
\multicolumn{1}{|l|}{decreased alertness}                       & \multicolumn{1}{l|}{interstitial lung disease}   & smoking                          \\ \hline
\multicolumn{1}{|l|}{diaphoresis}                           & \multicolumn{1}{l|}{left precordial pleuritic chest pain} & specific electrocardiogram changes \\ \hline
\multicolumn{1}{|l|}{diastolic blood pressure (mmHg)}           & \multicolumn{1}{l|}{leg swelling}                & spirometry (FVC)                 \\ \hline
\multicolumn{1}{|l|}{dimension of the jugular vein}         & \multicolumn{1}{l|}{lip cyanosis}                         & sudden onset of sharp chest pain   \\ \hline
\multicolumn{1}{|l|}{distant heart sounds}                      & \multicolumn{1}{l|}{low heart beat intensity}    & sweating                         \\ \hline
\multicolumn{1}{|l|}{dry cough}                                 & \multicolumn{1}{l|}{lung collapsed}              & systolic blood pressure (mmHg)   \\ \hline
\multicolumn{1}{|l|}{dyspnea}                                   & \multicolumn{1}{l|}{neck pain}                   & temperature                      \\ \hline
\multicolumn{1}{|l|}{electrical alternans on ekg}           & \multicolumn{1}{l|}{oxygen saturation (\%)}               & venous pressure                    \\ \hline
\multicolumn{1}{|l|}{fatigue}                                   & \multicolumn{1}{l|}{pain in the shoulders}       & volume overload                  \\ \hline
\multicolumn{1}{|l|}{fever}                                     & \multicolumn{1}{l|}{palpitations}                & weakness                         \\ \hline
\multicolumn{1}{|l|}{fluid around the heart}                    & \multicolumn{1}{l|}{pericardial effusion}        & wheezing                         \\ \hline
\multicolumn{1}{|l|}{friction rub}                              & \multicolumn{1}{l|}{plaques}                     & worsening of pain by inspiration \\ \hline
\multicolumn{1}{|l|}{frothy sputum}                             & \multicolumn{1}{l|}{pleural effusion}            & worsening of pain by lying down  \\ \hline
\end{tabular}
}
\label{tab:all_features_augm}
\caption{\textcolor{blue}{List of all features identified by DALL-M}}
\end{table}

\end{document}